\documentclass[a4paper, reqno, 12pt]{amsart}

\usepackage[utf8]{inputenc}
\usepackage{graphicx}
\usepackage{booktabs}
\usepackage{array}
\usepackage{hyperref}
\usepackage{url}
\usepackage{xcolor}
\usepackage{fancyhdr}
\usepackage{geometry}
\usepackage{amssymb} 
\usepackage{amsmath}
\usepackage{geometry}       
\usepackage{amsfonts}
\usepackage{framed}

\newcolumntype{L}[1]{>{\raggedright\arraybackslash}m{#1}}
\newcolumntype{C}[1]{>{\centering\arraybackslash}m{#1}}
\newcolumntype{R}[1]{>{\raggedleft\arraybackslash}m{#1}}

\usepackage{color}
\makeatletter

\@addtoreset{equation}{section}
\makeatother 

\theoremstyle{plain}
\newtheorem{theorem}{Theorem}[section]

\theoremstyle{definition}

\newtheorem{remark}[theorem]{Remark}

\usepackage{latexsym}
 
\title[The Shape of Consumer Behavior]{The Shape of Consumer Behavior:\\ A Symbolic and Topological Analysis of Time Series}

\author{Pola Bereta}
\address{Maastricht University, School of Business and Economics,
P.O.Box 616, 6200 MD, Maastricht,
The Netherlands.}
\email{p.bereta@student.maastrichtuniversity.nl}

\author{Ioannis Diamantis}
\address{Department of Data Analytics and Digitalisation,
Maastricht University, School of Business and Economics,
P.O.Box 616, 6200 MD, Maastricht,
The Netherlands.}
\email{i.diamantis@maastrichtuniversity.nl}


\begin{document}

\begin{abstract}
Understanding temporal patterns in online search behavior is crucial for real-time marketing and trend forecasting. Google Trends offers a rich proxy for public interest, yet the high dimensionality and noise of its time-series data present challenges for effective clustering. This study evaluates three unsupervised clustering approaches, Symbolic Aggregate approXimation (SAX), enhanced SAX (eSAX), and Topological Data Analysis (TDA), applied to 20 Google Trends keywords representing major consumer categories. Our results show that while SAX and eSAX offer fast and interpretable clustering for stable time series, they struggle with volatility and complexity, often producing ambiguous ``catch-all'' clusters. TDA, by contrast, captures global structural features through persistent homology and achieves more balanced and meaningful groupings.

We conclude with practical guidance for using symbolic and topological methods in consumer analytics and suggest that hybrid approaches combining both perspectives hold strong potential for future applications.

\bigbreak 

\end{abstract}

\maketitle

\section{Introduction}

Understanding the dynamics of consumer attention is critical for anticipating market trends, optimizing digital marketing strategies, and interpreting public interest in emerging technologies. While much of the existing literature on consumer behavior focuses on final outcomes, such as purchases, churn, or fraud detection \cite{9,10,11,12}, less attention is paid to earlier stages in the decision-making process, particularly to how interest forms and fluctuates over time. Yet, public attention often serves as a precursor to consumer action, and its temporal evolution offers valuable insights into behavior and intent.

Digital search data, such as that provided by Google Trends, offers a rich and underutilized source of information for analyzing consumer attention. Search queries can be interpreted as expressions of interest, which, when aggregated over time, form behavioral time series. Despite the increasing availability of such data, relatively few studies have examined multi-topic attention patterns over time, particularly in the context of general consumer categories. Existing work has primarily focused on narrow domains such as epidemiology, public health, or financial markets \cite{7,8}.

This paper addresses this gap by applying three time-series clustering techniques, Symbolic Aggregate approXimation (SAX), Enhanced SAX (eSAX), and Topological Data Analysis (TDA), to a dataset of weekly Google Trends data for 20 keywords representing major consumer topics. These methods have seen applications in other domains, but their combined use to analyze public attention across diverse product categories remains largely unexplored.

The methodological motivation for this study is twofold. First, symbolic representations such as SAX and eSAX offer interpretable and computationally efficient tools for summarizing time series. Second, TDA provides a fundamentally different lens, focusing on the topological structure of data and capturing complex, nonlinear features that persist across multiple scales. This dual framework builds on recent work by Hobbelhagen et al. \cite{16}, who demonstrated the value of symbolic and topological methods in financial settings, and echoes the broader call by Erevelles et al. \cite{15} to use big data not just for validation, but for discovery.

\smallbreak

Accordingly, this study seeks to answer the following research question:

\begin{framed}
\textit{How does consumer attention evolve over time across products and topics, and how can symbolic and topological methods be applied to time series data to analyze it?}
\end{framed}

In addressing this question, we evaluate the interpretability and clustering performance of each method and provide practical insights for behavioral analytics. Our contribution lies in the comparative analysis of symbolic and topological methods on consumer search behavior, and in demonstrating the strengths and trade-offs each approach brings to time series analysis.

\smallbreak

The paper is organized as follows: \S~\ref{sec:formatting} provides a brief review of related work on symbolic and topological methods in time series analysis. \S~\ref{sec:EDA} presents the exploratory data analysis of the Google Trends dataset. \S~\ref{method} introduces the symbolic approaches SAX and eSAX, while \S~\ref{sec:tda} discusses the application of Topological Data Analysis (TDA). Finally, \S~\ref{sec:discussion} offers a comparative discussion of the results, highlights limitations, and outlines directions for future research.

\section{Literature Review}\label{sec:formatting}

Understanding and modeling consumer attention over time is a relatively underexplored area in the literature, particularly when it comes to analyzing patterns across diverse topics. While traditional time series methods have been widely applied in fields such as economics, epidemiology, and marketing, more recent approaches, such as symbolic representations and topological techniques, have only begun to gain traction. This section reviews prior research on symbolic time series clustering, including Symbolic Aggregate approXimation (SAX) and its extensions, as well as recent developments in Topological Data Analysis (TDA) and their applications.

\subsection{SAX}
\label{sec:sax}

Symbolic Aggregate Approximation (SAX), introduced by Lin et al.\cite{1}, is a method for converting time series data into a sequence of symbols such as letters of an alphabet. The standard process of SAX can be described as follows:

\begin{itemize}
    \item[i.] Firstly, the data is normalized using Z-normalization ensuring that the input time series conforms to a standard normal distribution. This is critical because the SAX breakpoints are defined based on the Gaussian distribution’s quantiles.   
    \item[ii.] Then, dimensionality reduction is performed using {\it Piecewise Aggregate Approximation} (PAA). Roughly speaking, PAA divides the time series into equal-length segments and replaces its segment by its average value, while preserving the overall shape of the data.
    \item[iii.] Next, each average is mapped to a symbol, i.e. a letter of an alphabet, based on its relative position within the value distribution of the time series.
    \item[iv.] This results in a string of symbols (or in our case letters such as \textit{aabbcda}), which summarizes the shape of the time series symbolic form. These letters form the {\it SAX words}.
\end{itemize}

\begin{figure}[h!]
    \centering
    \includegraphics[width=1\linewidth]{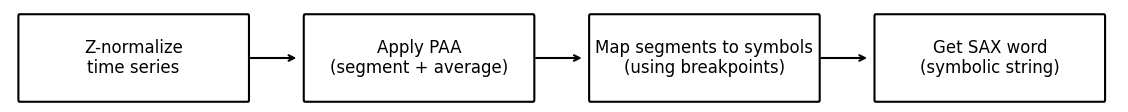}
    \caption{Flowchart of the SAX Algorithm}
    \label{fig:17}
\end{figure}

The above are summarized in Figure~\ref{fig:17}. 

\begin{figure}[h!]
    \centering
    \includegraphics[width=0.75\linewidth]{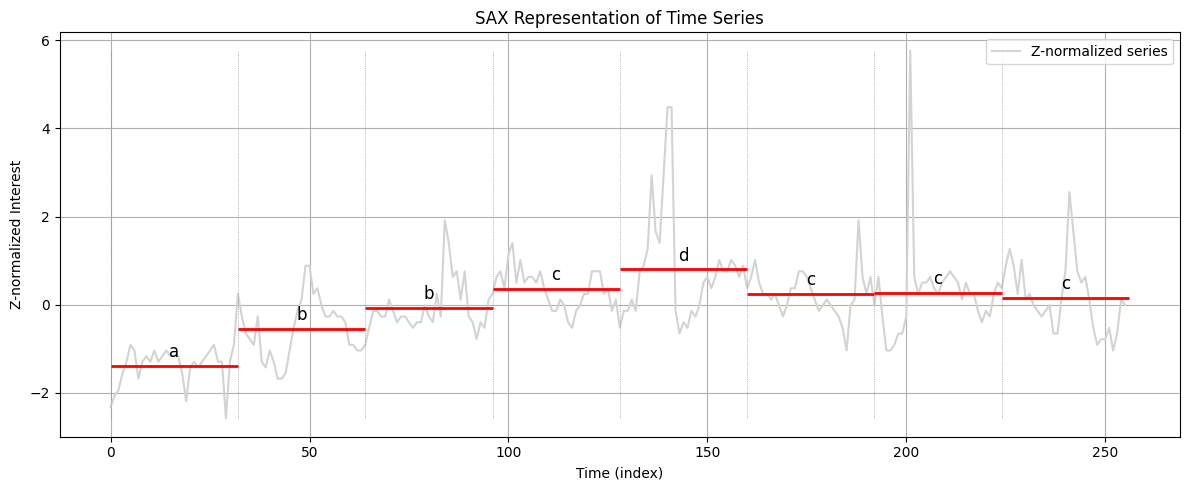}
    \caption{Example of SAX Output on Z-normalized Time Series}
    \label{fig:22}
\end{figure}

A detailed and intuitive illustration of the SAX procedure on one of the time series on our dataset is shown in Figure~\ref{fig:22}. The SAX word obtained in this example is \textit{abbcdccc}.

SAX has been used by Lin et al.\cite{1} for the analysis of time series with implications for streaming algorithms. In particular, they show how effective SAX’s approximation of the data is for clustering, classification, indexing and anomaly detection. The results of their analysis are that SAX’s symbolic representation can compete with other representations of the data or, in some cases, it is better than other methods depending on the nature of the data. 

\smallbreak

Aghabozorgi et al.\cite{2} describe SAX’s usage in applications such as pattern discovery and financial market analysis. They emphasize that SAX enables fast and scalable clustering by converting continuous time series into symbolic strings, which can then be indexed and compared efficiently. While the authors do not report new empirical evaluations of SAX themselves, they highlight its value in scenarios where dimensionality reduction and speed are critical. In particular, they point to studies where SAX has successfully uncovered hidden patterns in stock price movements and consumer transaction records, showing its potential in identifying structural similarities in noisy, high-dimensional data.

\smallbreak

When it comes to combining SAX with other methods, the original paper by Lin et al.\cite{1} demonstrates how SAX can be used with K-Means for motif discovery. Keogh et al.\cite{3} later applied K-Means to symbolic time series representations for classification and anomaly detection, which performed well, when compared to other nad more complex techniques.

\smallbreak

These results demonstrate the effectiveness of SAX in simplifying time series, as well as its adaptability and compatibility with other methods. Over time, several extensions and improvements have been proposed to enhance the performance of classical SAX. One such enhancement, eSAX, is described in the following subsection.


\subsection{Extended SAX (e-SAX)}\label{sec:cap-num}
\begin{figure}[h!]
    \centering
    \includegraphics[width=1\linewidth]{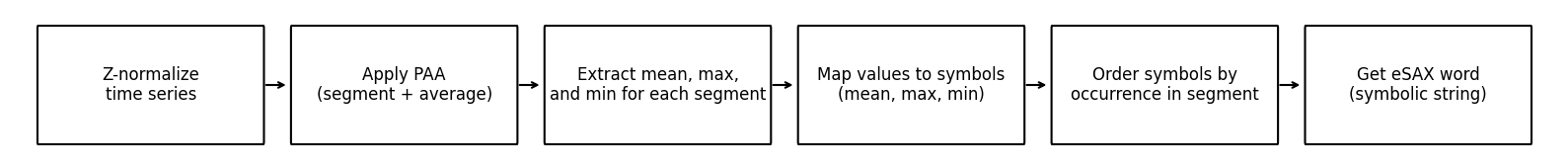}
    \caption{Flowchart of eSAX Algorithm}
    \label{fig:18}
\end{figure}

As mentioned above, SAX assigns symbols to the data based on the mean value of each segment. While this approach is effective for reducing dimensionality, it tends to ignore extreme values. As a result, sudden spikes or drops in the data may be averaged out, leading to the loss of important information. To address this limitation, an extension to the original model was proposed by Lkhagva et al.\cite{4} and later applied by the author in financial analysis. This method, known as {\it Extended Symbolic Aggregate approXimation} (eSAX), incorporates both the average and the variance of the data by considering the sequence of maximum, minimum, and mean values within each segment. This process is illustrated in Figure\ref{fig:18}. 

It is worth noting that eSAX was originally developed for financial time series data~\cite{4}. Furthermore, Hobbelhagen et al.~\cite{16} combined eSAX with clustering techniques for stock market analysis. Their findings indicate that eSAX performs relatively well in capturing broad stock movement patterns. However, the authors identified two key limitations when using eSAX in conjunction with clustering: first, one of the clusters contained the majority of stocks (39 out of 60); second, the small distances between clusters made it difficult to distinguish which stocks belonged to which group. Despite these limitations, eSAX was still able to group stocks exhibiting similar general trends. It effectively captured overarching patterns in stock behavior over time, though it lacked sensitivity to finer details and the precise timing of changes. According to their findings, eSAX provides a simple, interpretable approximation of time series data. However, for detecting more subtle differences between time series, the clustering results may be less informative.


\subsection{TDA}\label{sec:TDA}
Topological Data Analysis (TDA) is a collection of mathematical methods that can be used to analyse the ``shape'' of data. Unlike traditional statistical methods that focus on point-by-point analysis, TDA looks at how data points connect and form structures such as loops (or cycles), voids (or cavities) and higher-dimensional analogues (e.g. n-dimensional voids), across multiple scales. This makes TDA well-suited for analyzing noisy, high-dimensional datasets, which is a common characteristic for real-world time series, such as those used in this study.

\smallbreak

\begin{figure}[h!]
    \centering
    \includegraphics[width=1\linewidth]{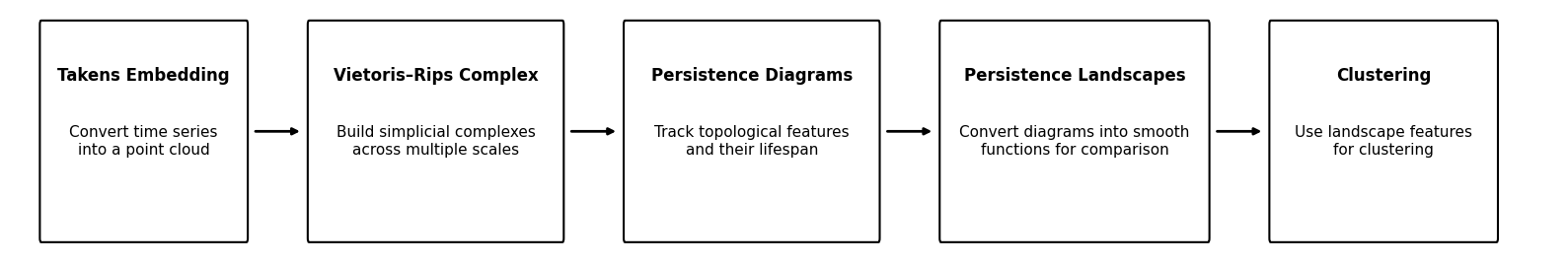}
    \caption{TDA Flowchart}
    \label{fig:19}
\end{figure}

One of the most important tools in TDA is {\it persistent homology}. This technique tracks how topological features, such as the previously mentioned structures, appear and disappear as we analyze the data across multiple scales. The goal is to identify the features that persist the longest, as these are typically the most meaningful and can offer valuable insights into the underlying structure of the data. At a high level, the process begins by constructing a family of {\it simplicial complexes} from the data at varying scales, and then computing the {\it birth and death} of topological features across these scales. This gives rise to {\it persistent diagrams}. A flowchart outlining the general steps of this procedure is provided in Figure~\ref{fig:19}. In Section~\ref{sec:tda} we describe the TDA steps in detail.

According to Chazal et al.~\cite{6}, TDA offers several advantages. Most notably, it does not assume linearity or any specific distribution in the data. Moreover, it is robust to noise and performs well on complex, real-world datasets. Due to these strengths, TDA has been applied successfully across a wide range of fields. More precisely, in astronomy, it has been used to analyze large-scale structures such as the cosmic web (Chen et al.~\cite{22}). In biology and medicine, applications include protein structure analysis (Kovacev-Nikolic et al.~\cite{24}) and the classification of histological images (Singh et al.~\cite{23}). In astrophysics and neuroscience, TDA has been employed to study shape-based features and spatial structures that are difficult to capture using traditional statistical or machine learning approaches (Taylor et al.~\cite{25}). For a survey on applications of TDA in various fields, the interested reader is referred to Wasserman et al.\cite{20}.

\smallbreak

Additionally, several studies have demonstrated that TDA outperforms traditional methods when applied to large and noisy datasets. For instance, Nicolau et al.~\cite{17} used a spectroscopic dataset of 4,000 bacterial samples to compare the performance of Principal Component Analysis (PCA), Hierarchical Clustering Analysis (HCA), and TDA. Their findings showed that TDA was able to distinguish between different bacterial groups, including small subgroups that PCA and HCA failed to identify. Notably, this was achieved without any data preprocessing, highlighting TDA’s robustness to noise. While preprocessing improved the performance of PCA and HCA slightly, TDA still outperformed both methods by uncovering meaningful and biologically relevant clusters. These results underscore one of TDA’s major strengths: its ability to detect intrinsic structure in complex, high-dimensional data.

\smallbreak

Singh et al.~\cite{18} demonstrated that TDA can be effectively used in a variety of domains beyond medicine. In their study, TDA was applied to three diverse datasets: gene expression profiles from cancer patients, voting records from the U.S. House of Representatives, and NBA player statistics. Once again, the results showed that TDA was able to uncover groups and patterns that were not detected by traditional methods such as PCA and hierarchical clustering. The use of TDA outside the biomedical context is particularly noteworthy. In the analysis of political data, TDA revealed patterns in inter-party cooperation over time, especially during major events such as the 2008 financial crisis. In sports analytics, it identified 13 distinct NBA playing styles, far exceeding the conventional classification into five player positions. These examples highlight TDA’s versatility and its effectiveness even when applied to relatively simple or low-dimensional datasets. These findings are consistent with the results presented in our study, further supporting the effectiveness of TDA in uncovering meaningful patterns in diverse datasets.

\smallbreak

Despite its growing success across various domains, TDA has yet to see widespread application in consumer behavior research. Nevertheless, it holds significant potential in this area, as consumer attention and behavior are highly dynamic and influenced by a wide range of factors. This paper therefore aims to explore how TDA, in combination with SAX and eSAX, can provide a novel perspective for analyzing noisy and heterogeneous time series data related to public interest.


\section{Exploratory Data Analysis (EDA)}
\label{sec:EDA}
\begin{figure}[h!]
    \centering
    \includegraphics[width=1\linewidth]{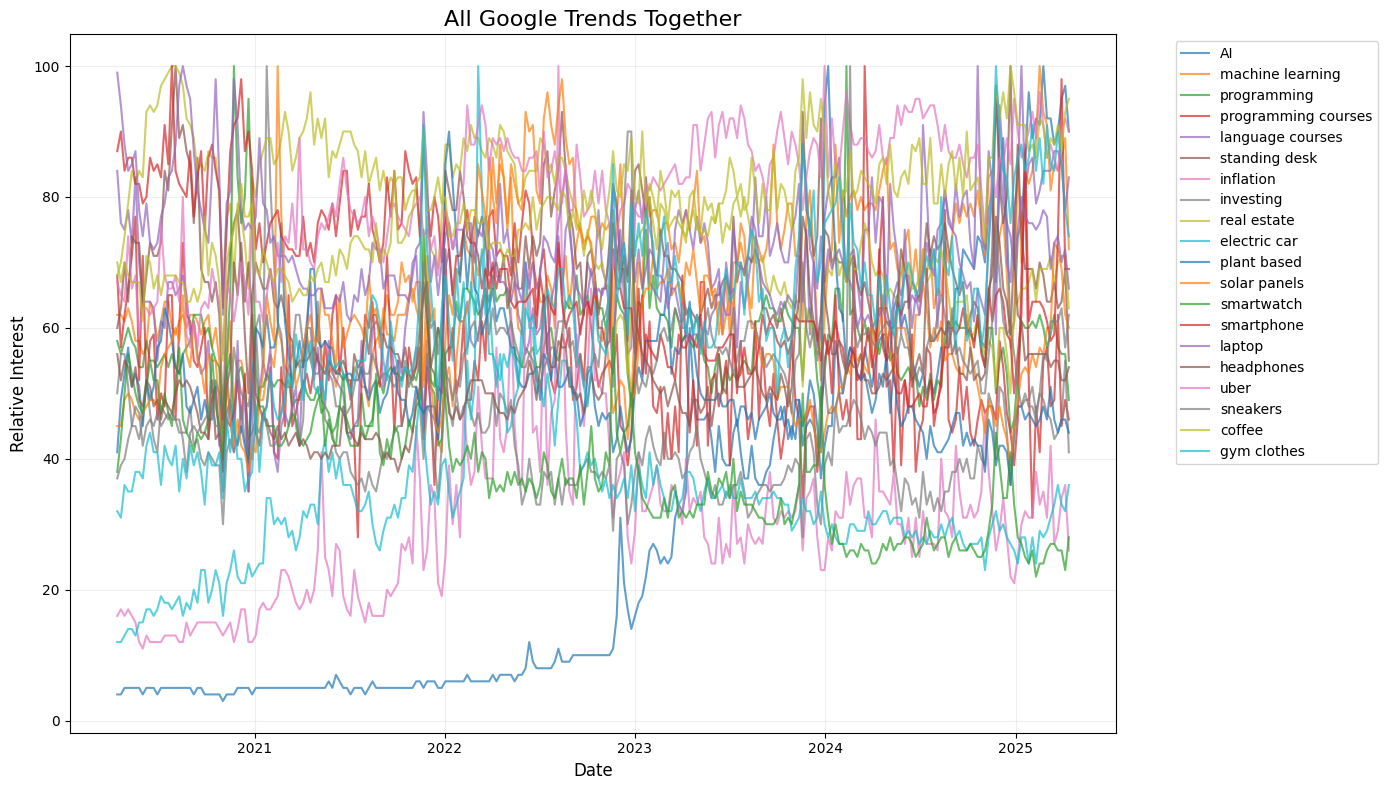}
    \caption{Google Trends Time Series}
    \label{fig:1}
\end{figure}

The data used in this study consist of time series representing weekly search interest in the United States. "Weekly search interest" refers to the number of Google searches for a given topic during each week. The data were obtained directly from Google Trends by downloading the time series for 20 selected keywords, including topics, products, and services. The observation period spans five years, from April 12, 2020, to April 13, 2025. Notably, the dataset contains no missing values. The raw time series data retrieved from Google Trends is illustrated in Figure~\ref{fig:1}.

A summary of key descriptive statistics for each keyword is provided in Table~\ref{tab:summary-stats}, including mean, standard deviation, minimum, and maximum values. This gives an initial sense of central tendency and variability across search terms.

\begin{table}[h!]
\centering
\scriptsize
\caption{Summary Statistics for Each Time Series (Keyword)}
\label{tab:summary-stats}
\begin{tabular}{L{2cm} C{1.1cm} C{1.1cm} C{1.1cm} C{1.1cm} C{1.1cm} C{1.1cm} C{1.1cm} C{1.1cm}}
\toprule
\textbf{Keyword} & \textbf{Count} & \textbf{Mean} & \textbf{Std} & \textbf{Min} & \textbf{25\%} & \textbf{50\%} & \textbf{75\%} & \textbf{Max} \\
\midrule
AI                  & 262 & 28.48 & 27.44 & 3   & 5   & 10   & 50.00 & 100 \\
coffee              & 262 & 78.08 & 7.89  & 55  & 72  & 78   & 82.75 & 100 \\
electric car        & 262 & 33.08 & 10.99 & 12  & 28  & 32   & 36    & 100 \\
gym clothes         & 262 & 61.10 & 13.76 & 31  & 53  & 62   & 69    & 100 \\
headphones          & 262 & 53.40 & 9.94  & 38  & 48  & 52   & 56    & 100 \\
inflation           & 262 & 29.75 & 12.34 & 11  & 20  & 30   & 36    & 100 \\
investing           & 262 & 42.44 & 8.00  & 28  & 37  & 41   & 46    & 100 \\
language courses    & 262 & 66.25 & 12.10 & 35  & 58  & 67   & 75    & 100 \\
laptop              & 262 & 71.29 & 8.36  & 57  & 65  & 70   & 75    & 100 \\
machine learning    & 262 & 67.70 & 10.84 & 40  & 60  & 68   & 76    & 100 \\
plant based         & 262 & 51.89 & 9.17  & 36  & 46  & 50   & 56    & 100 \\
programming         & 262 & 56.40 & 6.52  & 42  & 52  & 56   & 61    & 100 \\
programming courses & 262 & 53.66 & 10.38 & 28  & 46  & 52   & 59    & 100 \\
real estate         & 262 & 76.26 & 10.77 & 50  & 68  & 76   & 85    & 100 \\
smartphone          & 262 & 67.17 & 11.59 & 47  & 58  & 64   & 76    & 100 \\
smartwatch          & 262 & 38.85 & 11.64 & 22  & 31  & 38   & 44    & 100 \\
sneakers            & 262 & 55.09 & 7.80  & 35  & 51  & 55   & 59    & 100 \\
solar panels        & 262 & 59.82 & 12.16 & 38  & 51  & 58   & 65    & 100 \\
standing desk       & 262 & 67.30 & 8.91  & 46  & 62  & 67   & 72    & 100 \\
uber                & 262 & 80.73 & 9.50  & 56  & 75  & 83   & 88    & 100 \\
\bottomrule
\end{tabular}
\end{table}

For each search term, the data consist of a series of dates and corresponding search interest values. According to the Google News Initiative~\cite{21}, this data is aggregated from three platforms: Google Search, Google News, and YouTube. The interest values are normalized on a scale from 0 to 100 by Google Trends, where 100 represents the point of highest search popularity. As a result, the data reflect relative search interest over time rather than absolute search volumes within the United States.

\smallbreak

\begin{figure}[h!]
    \centering
    \includegraphics[width=1\linewidth]{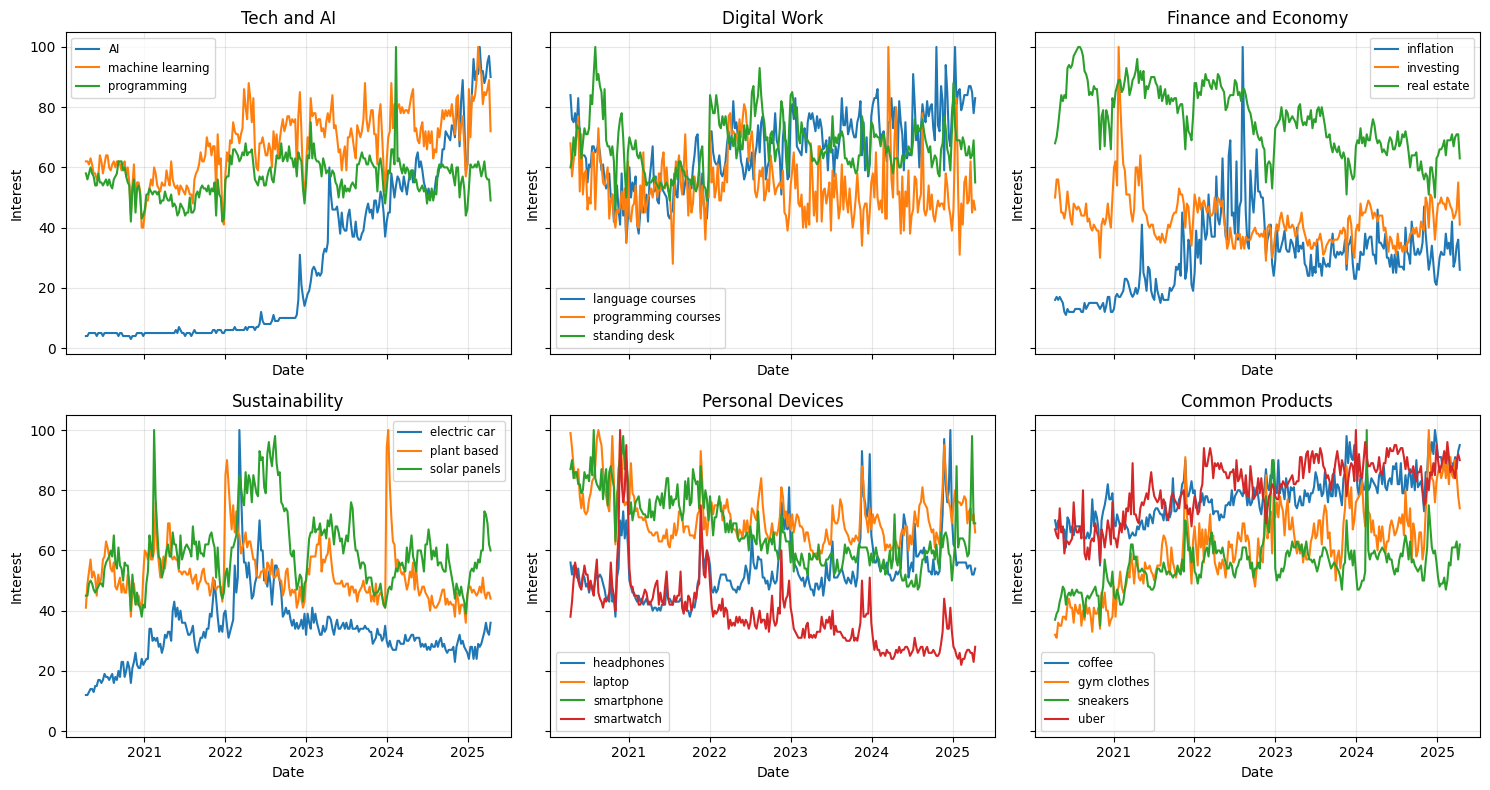}
    \caption{Time Series by Category}
    \label{fig:2}
\end{figure}

The 20 keywords were selected to reflect diverse aspects of modern consumer behavior and to offer insights into general public interest. The focus was placed on topics with long-term relevance, including those influenced by seasonal trends, major events such as the COVID-19 pandemic, and emerging technology. To facilitate interpretation, the keywords were organized into six thematic categories, each representing a common area of interest and containing a subset of representative time series (see Table~\ref{tab:keyword_categories} and Figure~\ref{fig:2}).

\begin{table}[h!]
\centering
\begin{tabular}{p{5cm} p{10cm}}
\toprule
\textbf{Category} & \textbf{Description} \\
\midrule
Tech \& AI & Topics that have gained attention in recent years\\
& because of technological development. \\
&\\
Digital Work & Changes in work culture, since remote work,\\
& online learning, and home office products\\
& became more relevant and common after COVID. \\
&\\
Finance \& Economy & Topics that directly impact consumer behavior,\\
& influenced by external events or market conditions. \\
&\\
Sustainability & Interest in environmentally conscious products\\
& due to growing relevance of green consumerism. \\
&\\
Personal Devices & Commonly used, everyday electronics\\
& that most people own and rely on. \\
&\\
Common Products & Items or services that did not clearly fit into the \\
& other categories but still represent popular topics\\
& of consumer attention. \\
\bottomrule
\end{tabular}
\caption{Overview of the Six Keyword Categories}
\label{tab:keyword_categories}
\end{table}

\begin{figure}[h!]
    \centering
    \includegraphics[width=0.75\linewidth]{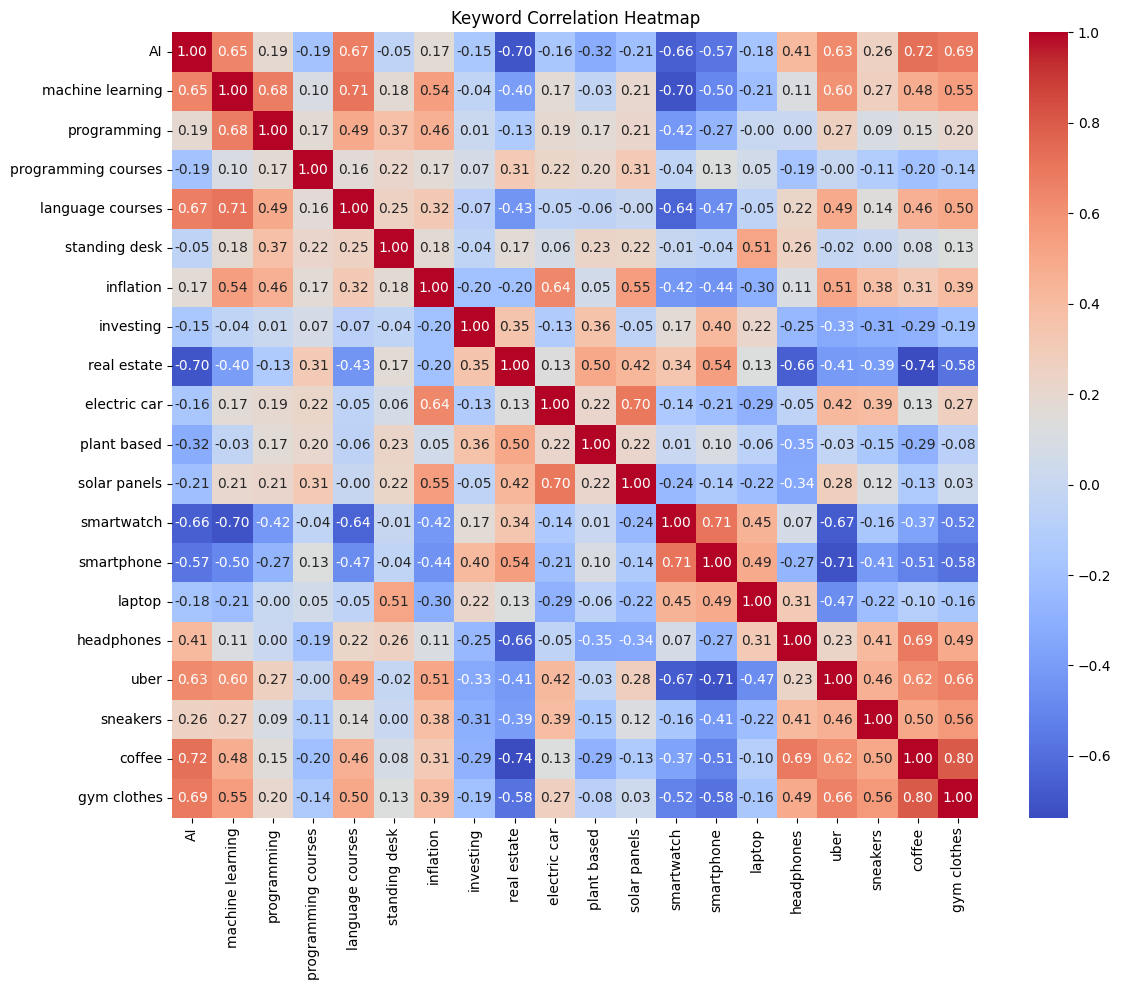}
    \caption{Correlation Table}
    \label{fig:correlations}
\end{figure}

To better understand the relationships between the selected keywords, a correlation matrix is presented in Figure~\ref{fig:correlations}. It captures both within- and cross-category relationships, with stronger correlations observed among keywords from similar domains. For example, \textit{coffee} and \textit{gym clothes} exhibit a strong positive correlation (0.8), as do \textit{AI} and \textit{machine learning} (0.65). In contrast, pairs like \textit{AI} and \textit{standing desk} (–0.05) show no meaningful correlation.

In addition to linear correlation, we also applied the Kolmogorov–Smirnov (K–S) test to assess distributional similarity between the keyword time series. The results (see Table~\ref{Ktest}) highlight several pairs with statistically significant differences, further supporting the heterogeneity of the selected keywords.

\begin{table}[h]\label{Ktest}
\centering
\caption{Kolmogorov–Smirnov Test Results for Normality (5\% significance level)}
\begin{tabular}{@{}lccc@{}}
\toprule
\textbf{Keyword} & \textbf{KS Statistic} & \textbf{p-value} & \textbf{Reject H$_0$ at 5\%} \\
\midrule
AI                    & 0.2611 & 0.0000 & Yes \\
electric car          & 0.1547 & 0.0000 & Yes \\
headphones            & 0.1678 & 0.0000 & Yes \\
plant based           & 0.1388 & 0.0001 & Yes \\
smartphone            & 0.1269 & 0.0004 & Yes \\
solar panels          & 0.1159 & 0.0016 & Yes \\
uber                  & 0.1134 & 0.0022 & Yes \\
sneakers              & 0.1096 & 0.0034 & Yes \\
programming courses   & 0.1051 & 0.0056 & Yes \\
laptop                & 0.0997 & 0.0101 & Yes \\
smartwatch            & 0.0965 & 0.0142 & Yes \\
investing             & 0.0941 & 0.0180 & Yes \\
inflation             & 0.0915 & 0.0234 & Yes \\
real estate           & 0.0818 & 0.0566 & No \\
standing desk         & 0.0681 & 0.1683 & No \\
machine learning      & 0.0610 & 0.2719 & No \\
programming           & 0.0603 & 0.2854 & No \\
coffee                & 0.0581 & 0.3270 & No \\
gym clothes           & 0.0547 & 0.3980 & No \\
language courses      & 0.0439 & 0.6780 & No \\
\bottomrule
\end{tabular}
\end{table}

In the following subsections, additional exploratory methods such as rolling averages, STL decomposition, and stationarity tests are applied to deepen our understanding of the temporal dynamics within the data. Section~\ref{method} then presents a more advanced analysis, where SAX, eSAX, and TDA are employed as modeling techniques.

\subsection{Standardization}
In this subsection, we describe the data preparation process for two of the methods used in this study: SAX and eSAX. As mentioned in the previous section, Google Trends data are min-max normalized, with 100 indicating peak interest and 0 representing the lowest or no interest at all. However, both SAX and eSAX require the time series to be Z-normalized. This is because SAX divides the value range into intervals based on breakpoints derived from the standard normal distribution~\cite{1}. In particular, if the input data remain unnormalized or only min-max normalized, the breakpoints would not align with the actual distribution of the data, potentially resulting in misleading symbolic representations. This issue becomes particularly pronounced when comparing time series with differing levels of volatility or flatness, as is the case with the Google Trends dataset. To address this, each time series is Z-normalized to have a mean of 0 and a standard deviation of 1, following the transformation:

\begin{equation}
z_t = \frac{x_t - \mu}{\sigma}
\end{equation}
where \( x_t \) is the original value at time \( t \), and \( \mu \) and \( \sigma \) denote the series mean and standard deviation, respectively.

\subsection{Rolling average and volatility bands}
\begin{figure}[ht]
    \centering
    \includegraphics[width=0.72\linewidth]{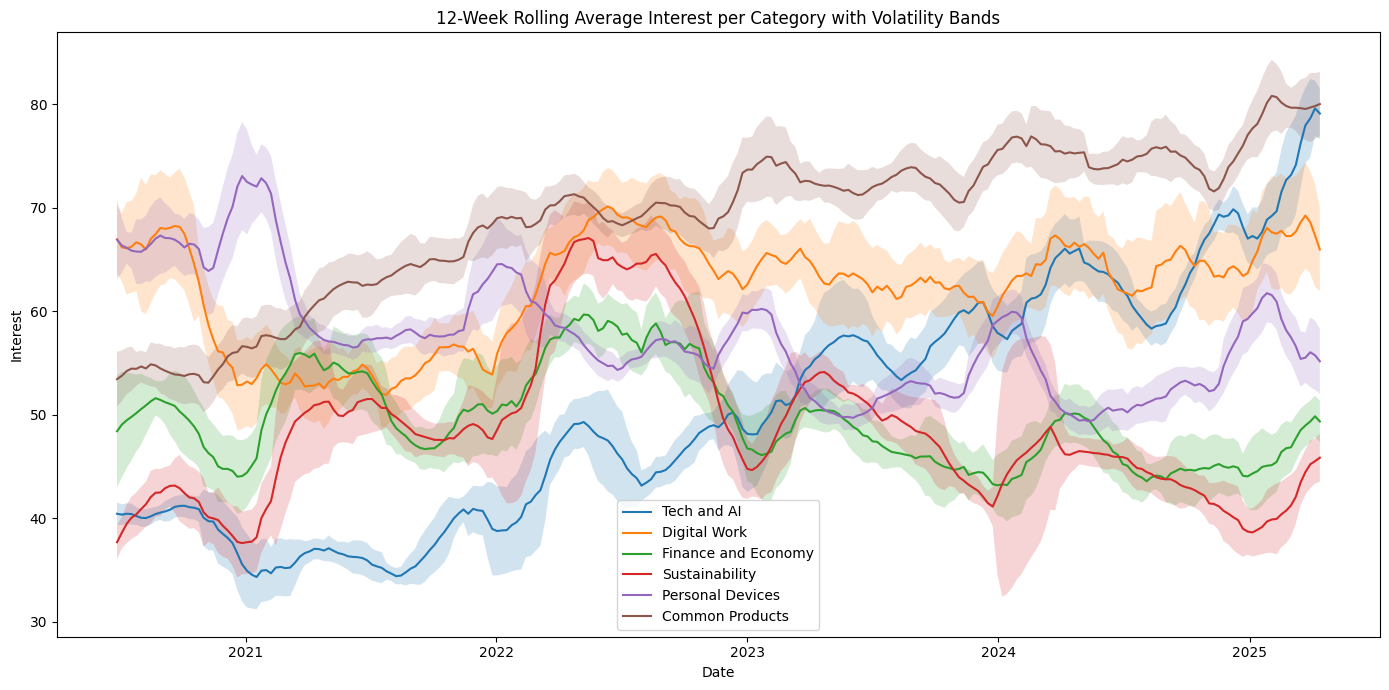}
    \caption{Rolling Average Interest per Category with Volatility Bands}
    \label{fig:3}
\end{figure}

To gain a better understanding of the data, we compute a 13-week rolling average along with corresponding \textit{volatility bands} for each category. Here, \textit{volatility} refers to the extent to which search interest for each keyword within the six categories fluctuates over time. At each time step, the average interest over the preceding 13 weeks is calculated, and the window is then shifted forward by one week. Since the data are recorded weekly, the 13-week window corresponds to approximately one quarter of a year. The outcome of this process is visualized in Figure~\ref{fig:3}, where the shaded regions (volatility bands) represent the rolling standard deviation around the moving average.

Figure~\ref{fig:3} illustrates distinct temporal patterns and volatilities across the six keyword categories. The "Common Products" category exhibits a steady upward trend with relatively stable volatility. The "Digital Work" category shows a decline following the pandemic, before gradually rising again, an interesting pattern that reflects a temporary drop in interest in remote work tools and skill development (around late 2022 to early 2023), followed by renewed engagement as hybrid and remote work models gained traction. The "Personal Devices" group displays clear seasonal spikes, particularly during year-end shopping periods. In contrast, "Finance and Economy" is marked by irregular spikes, reflecting the public's responsiveness to unpredictable external events. The "Sustainability" category begins with low interest but becomes increasingly volatile over time, especially in early 2024. Finally, the "Tech and AI" category transitions from a niche topic to one of the most prominent areas of interest by 2025.
These observations support the initial categorization, as each group demonstrates unique temporal dynamics and reflects different aspects of consumer behavior and public attention.

\subsection{Descriptive statistics }
\begin{figure}[h!]
    \centering
    \includegraphics[width=1\linewidth]{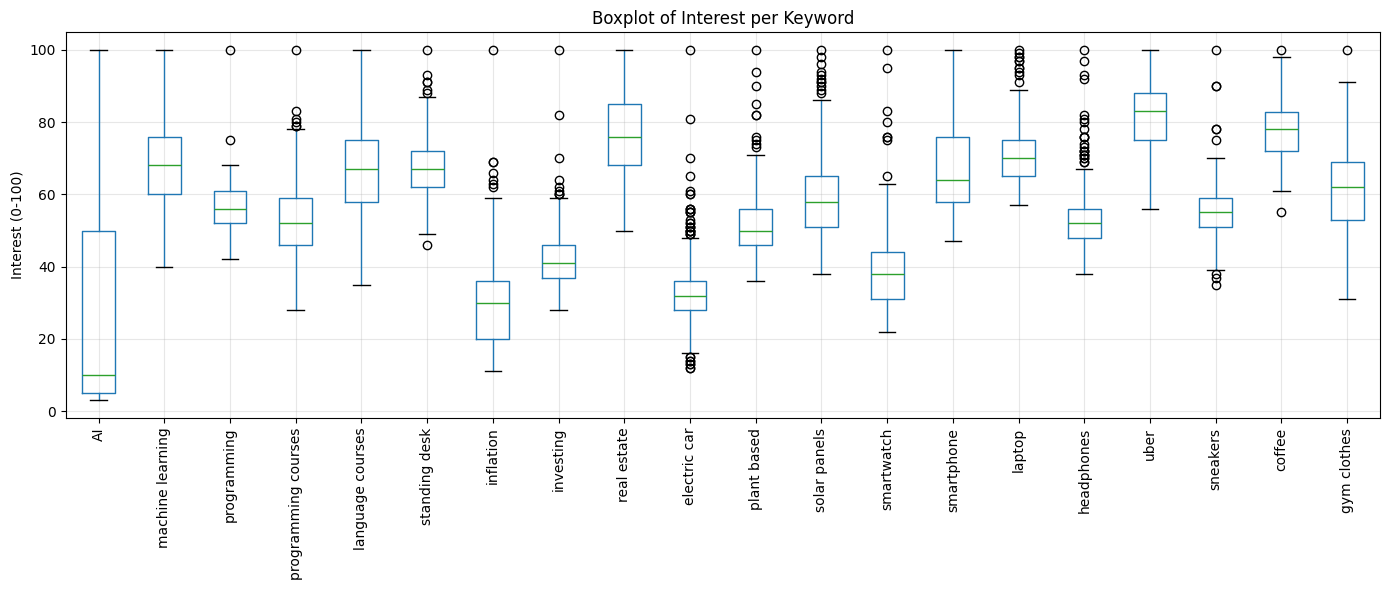}
    \caption{Boxplots of Interest per Keyword}
    \label{fig:4}
\end{figure}

In this subsection, we focus on the individual keywords within the above-mentioned 6 categories. This is done in order to analyze how time series reflect the distribution of the search interest. Recall that the dataset contains 20 keywords in total; each represented as a weekly time series over five years (262 data points per keyword) with no missing values. In Figure~\ref{fig:4} we present a Boxplot per keyword  to illustrate the distribution of interest within their respective categories.

\noindent Some remarks are now due.

\begin{remark}\rm
\,
\begin{itemize}
    \item[i.] Keyword \textit{AI} within the Tech \& AI category has the highest volatility. It has extreme outliers and a wide spread, which confirms the results from the previous section (recall Figure~\ref{fig:3}).
    \item[ii.] The keywords \textit{inflation} and \textit{electric car} have low medians and many outliers. This shows that the search interest is influenced by external events (for example, policy changes or politics in general).
    \item[iii.] One observation that is visible across almost all keywords is that there are many outliers close to 100. These are the spikes in attention that are expected in public interest that rise unexpectedly because of trends, news or events.
\end{itemize}
\end{remark}

\begin{figure}[h!]
    \centering
    \includegraphics[width=0.85\linewidth]{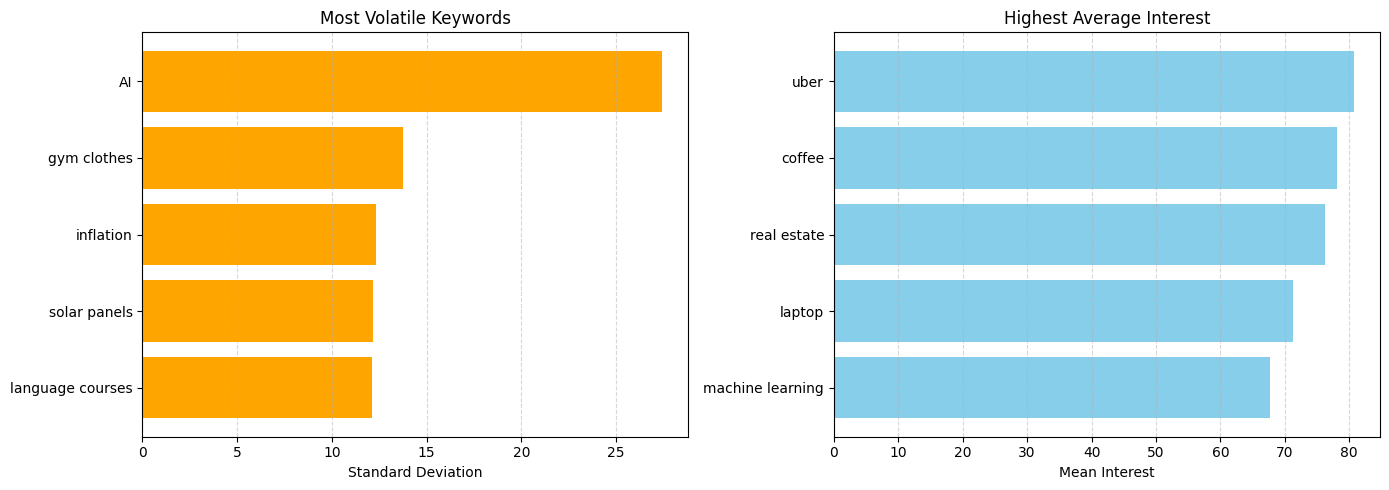}
    \caption{Most Volatile and Highest Interest Keywords}
    \label{fig:5}
\end{figure}

Figure~\ref{fig:5} illustrates the top five keywords for both volatility and the highest average interest. 
The most volatile keywords after \textit{AI} are \textit{gym clothes} and \textit{inflation}. The most important ones, in terms of highest average interest, are \textit{uber, coffee} and {\it real estate}. This does not come as a surprise: \textit{uber} is a very popular service, and \textit{AI} is volatile by definition, due to recent developments. The purpose of Figure~\ref{fig:5} is to demonstrate that the selected terms are highly relevant, reflecting what people care about, what they engage with, and how their attention shifts in response to changes in the world around them.


\subsection{Seasonal-Trend Decomposition and Augmented Dickey–Fuller test}
Moving beyond summary statistics, Seasonal-Trend decomposition using LOESS, abbreviated as STL, is applied to the data. Originally introduced by Cleveland et al.~\cite{5}, STL decomposes a time series into three components: trend, seasonality, and residuals. This method is used to uncover the underlying structure of the data by separating it into interpretable parts. It provides an initial glimpse into the complexity of each series before any transformations are applied or models are built. Unlike classical decomposition methods, STL can handle any type of seasonality, allows for missing values, and is resistant to outliers due to its use of LOESS (locally weighted regression) for smoothing. This makes it particularly well-suited for real-world time series data, such as the weekly search interest examined in this study.

\smallbreak

\begin{figure}[h!]
    \centering
    \includegraphics[width=0.7\linewidth]{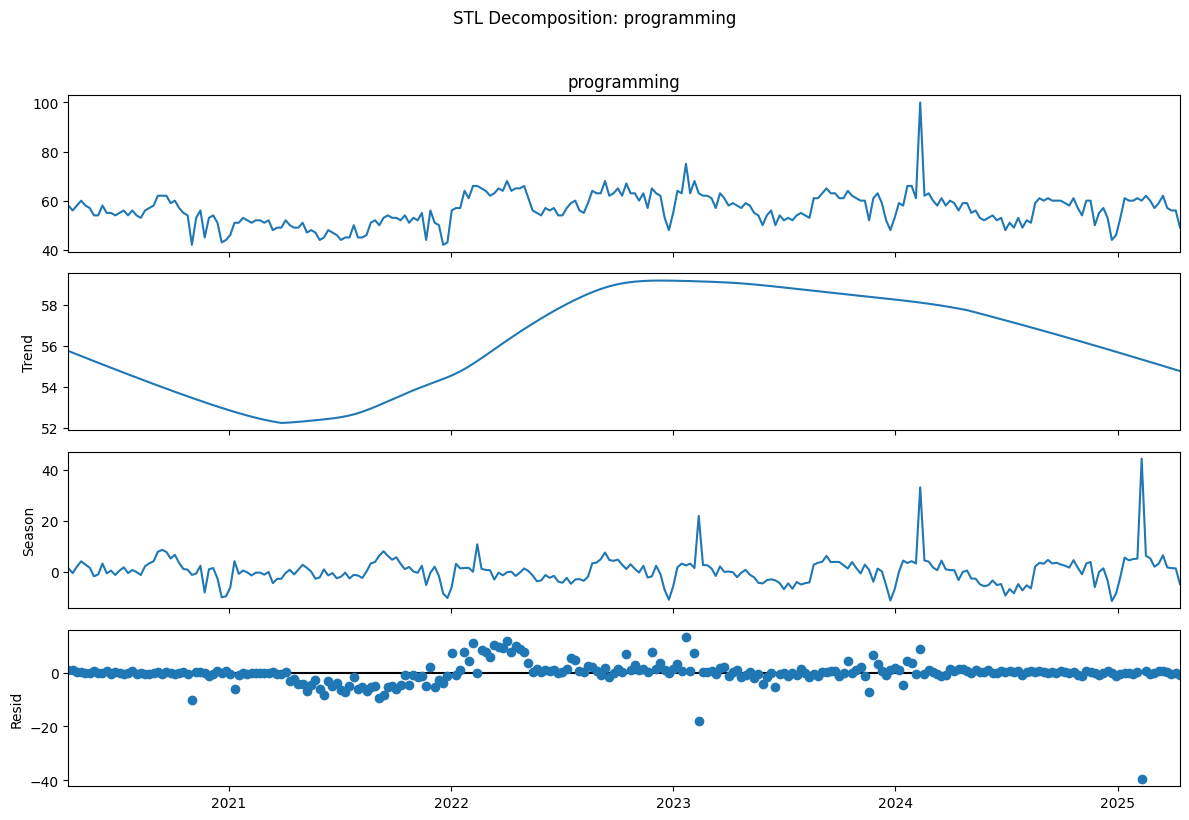}
    \caption{Seasonal Decomposition of the keyword \textit{programming}}
    \label{fig:6}
\end{figure}

STL is applied and plotted for each of the keywords (see Figure~\ref{fig:6} for an example). The strong seasonality observed in the time series (as previously shown in Figure~\ref{fig:5}) is effectively captured by the seasonal component of the STL decomposition, illustrated in the third panel of Figure~\ref{fig:6}.

Further analysis of the stationarity of the time series is conducted as a complementary step to STL decomposition, in order to extract deeper insights into the data's behavior. Stationarity refers to the property of a time series whose statistical characteristics, such as mean and variance, remain constant over time. To assess stationarity, the Augmented Dickey–Fuller (ADF) test is applied. This test evaluates the presence of a unit root, which indicates a stochastic trend that drives the series away from its mean, rendering it {\it non-stationary}. The ADF test is framed as a hypothesis test with the following null and alternative hypotheses:

\begin{center}
\begin{framed}
$H_0$:\ {\rm series has a unit root} \ \ \ \ \ \  vs \ \ \ \ \ \  $H_1$:\ {\rm series does not have a unit root}
\end{framed}
\end{center}

The null hypothesis is rejected if the p-value is below the significance level of 0.05. Furthermore, the more negative the ADF test statistic, the stronger the evidence against the presence of a unit root, thereby supporting the rejection of the null hypothesis~\cite{26}.

\begin{figure}[h!]
    \centering
    \includegraphics[width=0.75\linewidth]{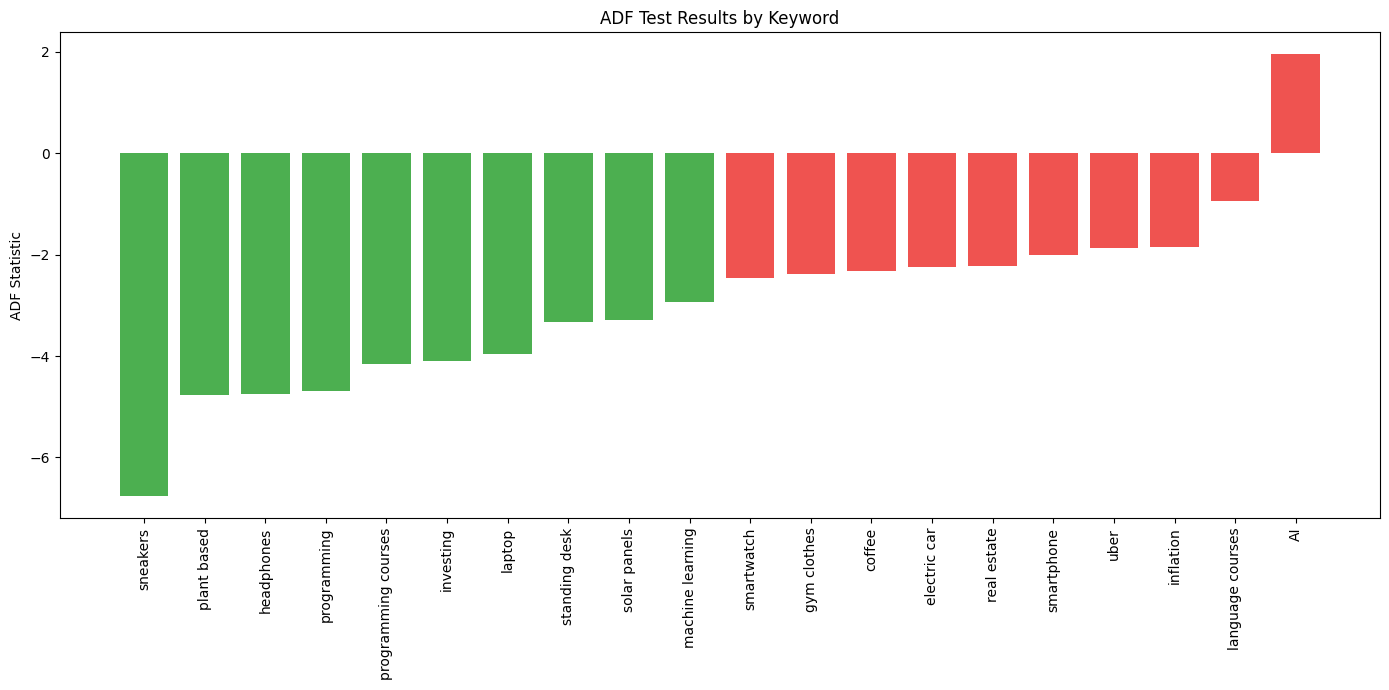}
    \caption{ADF Test Results}
    \label{fig:7}
\end{figure}

\noindent The results, presented in Figure~\ref{fig:7}, show that out of 20 keywords, exactly half exhibit stationarity at the 5\% significance level, while the remaining 10 are non-stationary. Green bars (e.g., \textit{sneakers}, \textit{plant based}, \textit{headphones}, \textit{programming}) represent stationary time series with low p-values and strongly negative ADF test statistics. In contrast, red bars (e.g., \textit{AI}, \textit{language courses}, \textit{inflation}, \textit{uber}) indicate non-stationary series, with p-values above 0.05. A detailed overview of the ADF test statistics and corresponding p-values is provided in Table~\ref{tab:adf-results}.

\begin{table}[h!]
\centering
\caption{ADF Test Results per Keyword}
\scriptsize
\label{tab:adf-results}
\begin{tabular}{L{4cm} C{2.5cm} C{2.5cm} C{2.5cm}}
\toprule
\textbf{Keyword} & \textbf{ADF Statistic} & \textbf{p-value} & \textbf{Stationary} \\
\midrule
sneakers            & -6.7582 & 0.0000 & Yes \\
plant based         & -4.7622 & 0.0001 & Yes \\
headphones          & -4.7484 & 0.0001 & Yes \\
programming         & -4.6832 & 0.0001 & Yes \\
programming courses & -4.1599 & 0.0008 & Yes \\
investing           & -4.0949 & 0.0010 & Yes \\
laptop              & -3.9522 & 0.0017 & Yes \\
standing desk       & -3.3184 & 0.0141 & Yes \\
solar panels        & -3.2960 & 0.0151 & Yes \\
machine learning    & -2.9353 & 0.0414 & Yes \\
smartwatch          & -2.4646 & 0.1243 & No \\
gym clothes         & -2.3711 & 0.1501 & No \\
coffee              & -2.3285 & 0.1629 & No \\
electric car        & -2.2486 & 0.1891 & No \\
real estate         & -2.2284 & 0.1961 & No \\
smartphone          & -2.0090 & 0.2826 & No \\
uber                & -1.8677 & 0.3474 & No \\
inflation           & -1.8521 & 0.3549 & No \\
language courses    & -0.9427 & 0.7736 & No \\
AI                  &  1.9524 & 0.9986 & No \\
\bottomrule
\end{tabular}
\end{table}

Although STL decomposition is not used as a preprocessing step for the modeling techniques in this study, it plays an important exploratory role. By decomposing each time series into interpretable components (trend, seasonality and residual), STL offers valuable insight into the underlying temporal dynamics. This structural understanding supports the decision to apply methods such as SAX, eSAX, and TDA, which are designed to capture complex patterns in the data without imposing strong statistical assumptions. While non-stationary time series can be transformed using various techniques to enforce stationarity~\cite{13}, such preprocessing may inadvertently remove meaningful or informative structure~\cite{14}. Therefore, stationarity is not enforced in this study. Z-normalization is sufficient for SAX and eSAX, and TDA is inherently robust to raw, noisy, and non-stationary data. 


\section{Symbolic Methods}
\label{method}
We are now in a position to apply SAX and eSAX, to the dataset and conduct a comparative analysis of their results. Recall that SAX and eSAX simplify time series by transforming them into symbolic representations, thereby reducing dimensionality while preserving essential patterns.

\subsection{SAX}
The original SAX method proposed by Lin et al.\cite{1} has been adapted for this analysis by incorporating a {\it sliding window} approach. This modification allows for the capture of more localized and granular patterns within the time series, as demonstrated by Zhang et al.\cite{19}. For convenience, we refer to this modified version sinmply as SAX.

\smallbreak

\begin{figure}[h!]
    \centering
    \includegraphics[width=1\linewidth]{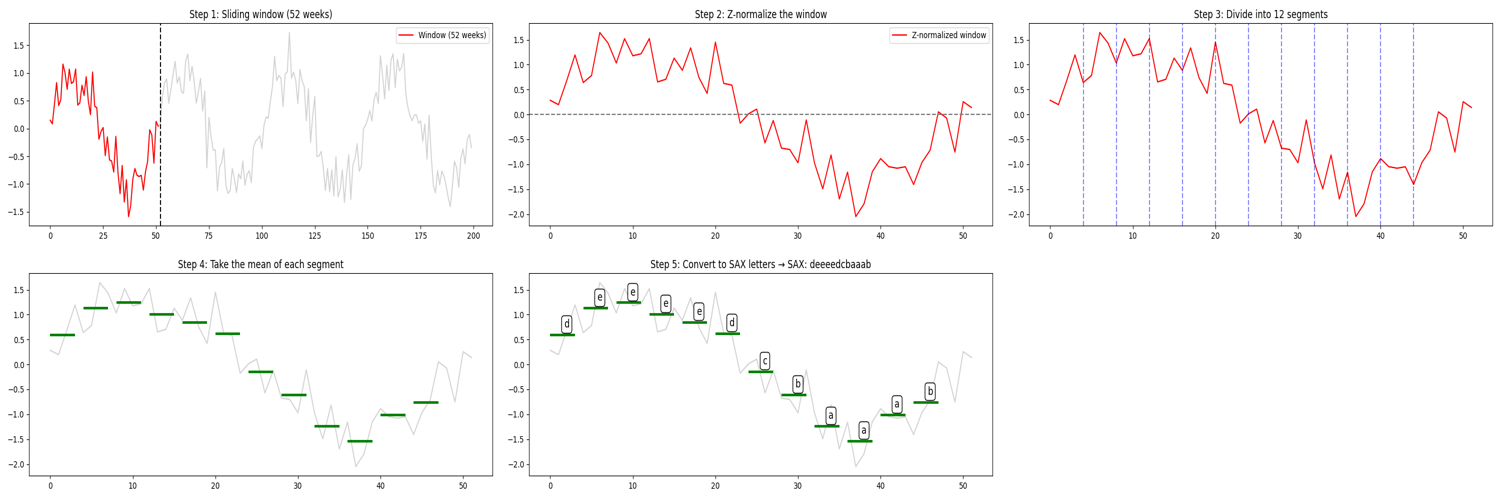}
    \caption{Sliding Window SAX Process}
    \label{fig:8}
\end{figure}

The process, illustrated in Figure~\ref{fig:8}, begins by applying a sliding window of size 52 data points, corresponding to approximately one year of weekly Google Trends data. Within each window, Z-normalization is first applied to ensure standardization. The normalized data is then divided into 12 equal-length segments, roughly corresponding to monthly intervals. The mean value of each segment is calculated, and each segment is assigned a symbol from a predefined alphabet based on its relative value. The window then advances by one time step, and the process is repeated. This produces a sequence of 211 SAX words for each time series, as determined by the following formula:
\[
\text{Number of SAX Words} = \text{Series Length} - \text{Window Size} + 1 = 262 - 52 + 1 = 211
\]

\begin{figure}[h!]
    \centering
    \includegraphics[width=1\linewidth]{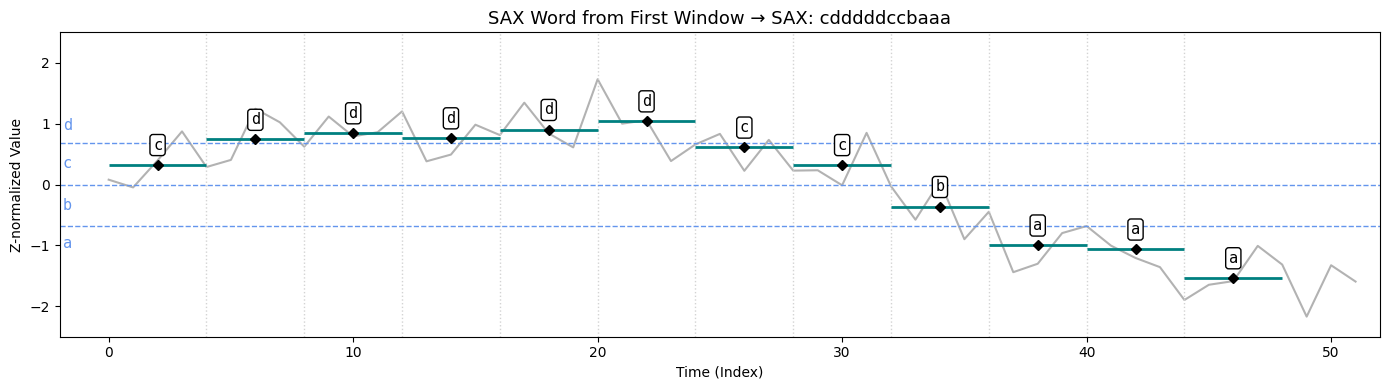}
    \caption{Example of SAX word for the first window frame of \textit{programming} keyword}
    \label{fig:SAX_WORD_EXAMPLE}
\end{figure}

Figure~\ref{fig:SAX_WORD_EXAMPLE} illustrates how the first SAX word is generated from the initial sliding window. Once the algorithm extracts all SAX words from each window, the resulting symbolic representation can be used for downstream analysis, specifically clustering, which is discussed in Section~\ref{kmeanscl}.

\subsubsection{K-means Clustering}\label{kmeanscl}
\begin{figure}[h!]
    \centering
    \includegraphics[width=0.4\linewidth]{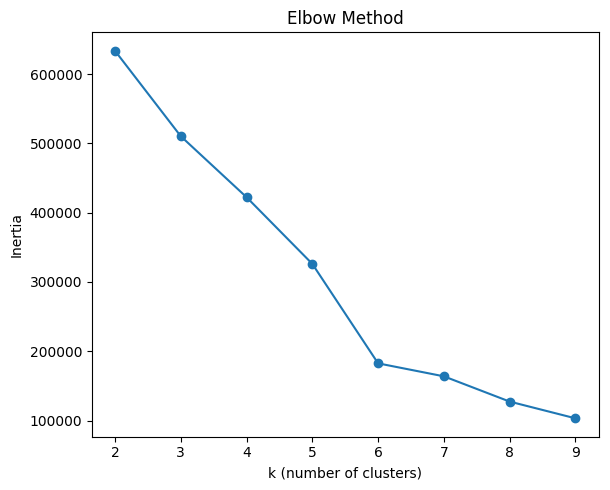}
    \caption{Elbow Method}
    \label{fig:elbow}
\end{figure}

K-means clustering was applied to the SAX output, with the number of clusters set to six. This choice was guided by the elbow method, which indicated an optimal cluster count at $k=6$ (see Figure~\ref{fig:elbow}).

The algorithm assigns each data point to one of the clusters based on its proximity to the cluster centers. The resulting clustering, shown in Figure~\ref{fig:9}, reveals a mixed performance. While some clusters appear inconsistent (or even somewhat ``chaotic''), others exhibit clear and coherent groupings. Clusters 1 through 4, along with Cluster 6, are well-structured, each containing a small number of time series that share similar patterns. In contrast, Cluster 5 appears to function as a ``catch-all'' group, consisting of six diverse time series that do not clearly fit into any of the other clusters.

\smallbreak

In order to test the performance of K-means, the silhouette score was used, which measures how similar an object is to its own cluster compared to other clusters. Specifically, the silhouette score ranges from $-1$ to $1$, where higher values mean that the data point is well grouped in its cluster, and not closely related to the neighbouring clusters~\cite{27}. It is defined as follows:
\[
s = \frac{b - a}{\max(a, b)}.
\]
where $a$ is the average distance between a point and all other points in the same cluster, and $b$ is the average distance between the point and all points in the nearest neighbouring cluster. The overall silhouette score is calculated as the average of the individual scores across all data points, offering a summary measure of clustering quality.

In our case, the silhouette score is 0.320 (see Table~\ref{tab:silhouette-scores}), indicating a weak-to-moderate clustering structure. While meaningful clusters do exist, the boundaries between them are relatively unclear. This suggests that SAX successfully captures and groups time series with simple, well-defined patterns but tends to oversimplify more complex series, merging them into a single, less coherent cluster (Cluster 5). These results raise the question of whether an alternative approach could yield more precise or informative groupings. To check whether the ``chaotic'' clusters can be broken down into more meaningful groups or not, in \S~\ref{hierclu} a different clustering approach is applied.

\subsubsection{Hierarchical Clustering}\label{hierclu}
Hierarchical Clustering is now applied to the SAX output. The Ward method~\cite{28} of clustering has been used to control how the distances between clusters are calculated during the merging process. The clustering starts by treating each keyword as its own cluster, then, with iterations, it calculates distances between all clusters using the Euclidean distance. Finally, it merges the two closest clusters together, and this process is repeated until 6 clusters are created. Six clusters were chosen for consistency (recall Figure~\ref{fig:elbow}).

\smallbreak

\begin{figure}[h!]
  \centering

  \begin{minipage}[b]{0.48\linewidth}
    \centering
    \includegraphics[width=\linewidth]{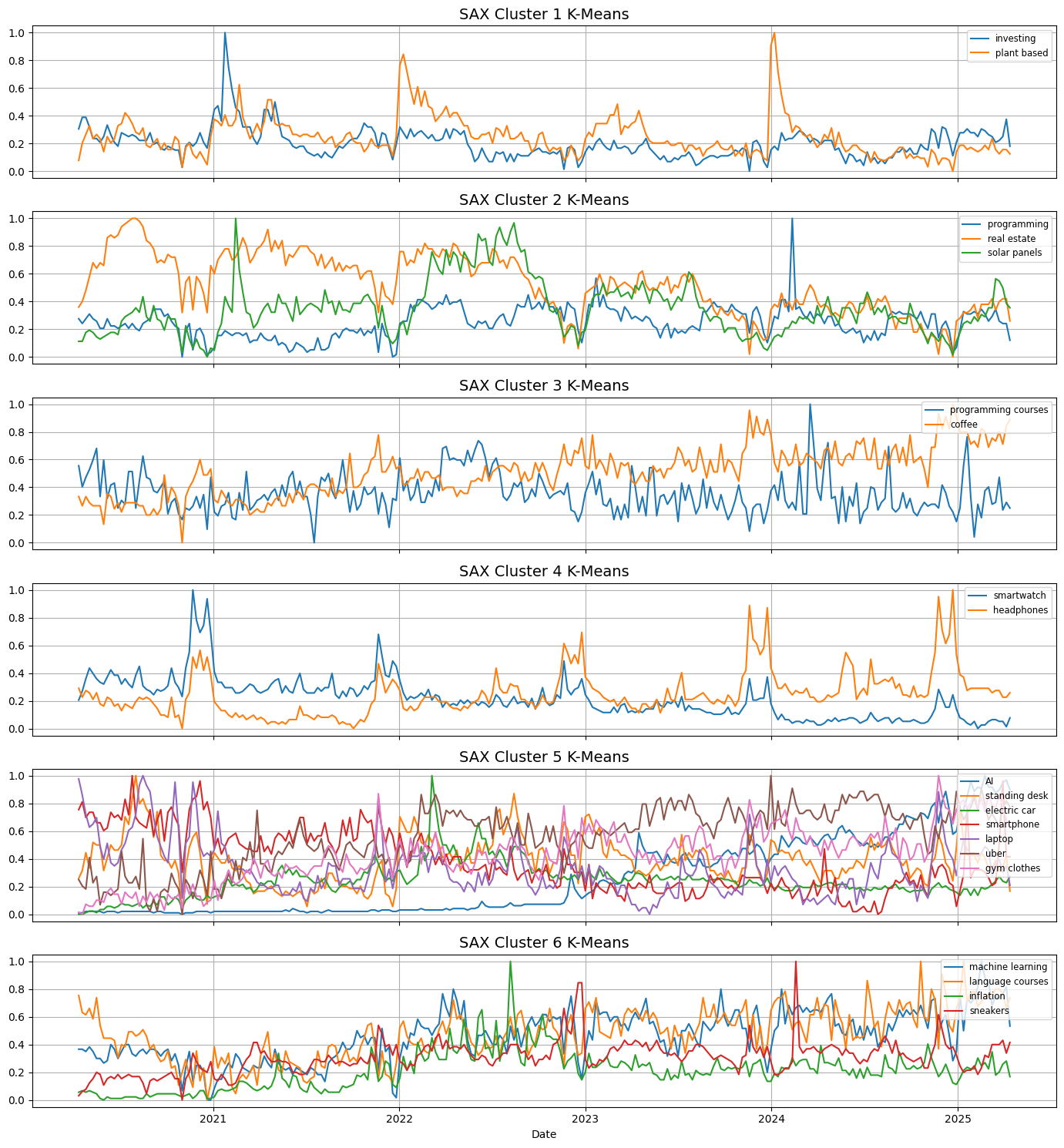}
    \caption{SAX K-Means Clustering}
    \label{fig:9}
  \end{minipage}
  \hfill
  \begin{minipage}[b]{0.48\linewidth}
    \centering
    \includegraphics[width=\linewidth]{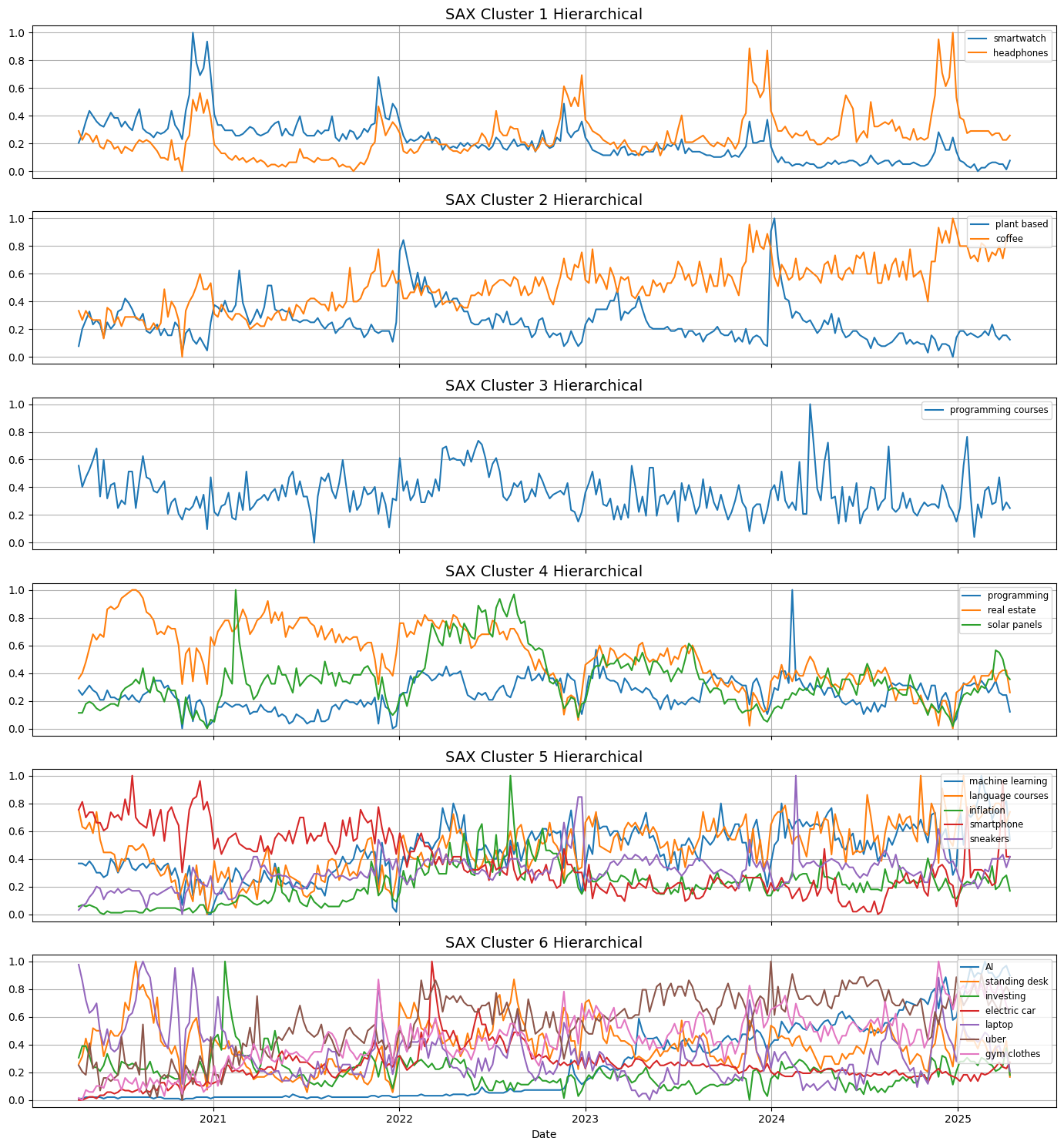}
    \caption{SAX Hierarchical Clustering}
    \label{fig:10}
  \end{minipage}

\end{figure}

The resulting clusters, illustrated in Figure~\ref{fig:10}, differ slightly from those produced by K-means. These differences arise from the nature of the Ward method, which builds clusters gradually through bottom-up merging. While some clusters remain consistent across both methods, including several that match exactly, others show structural variations. Once again, Clusters 1 through 4 exhibit coherent groupings. Cluster 5 appears more loosely defined, combining time series with somewhat similar trends. In this case, Cluster 6 includes the most volatile and unpredictable series that do not align well with any other group. The silhouette score for this clustering is 0.355, representing a modest improvement over the previous result and suggesting moderate clustering quality.

Since the silhouette score obtained from hierarchical clustering is not substantially higher than that of K-means, we explore whether clustering performance can be improved using an alternative representation. To this end, we consider an extension of the SAX algorithm known as \textit{enhanced SAX} (eSAX). The process is described in the following subsection.

\subsection{eSAX}

eSAX is an extension of the classical SAX method that incorporates additional information from each segment within a sliding window. While SAX summarizes each segment using only its mean, eSAX enhances this representation by also including the minimum and maximum values. By retaining these extreme values, previously averaged out in the original method, eSAX captures more detail within each segment, resulting in a symbolic representation that contains three times more information. 

\smallbreak

\begin{figure}[h!]
    \centering
    \includegraphics[width=1\linewidth]{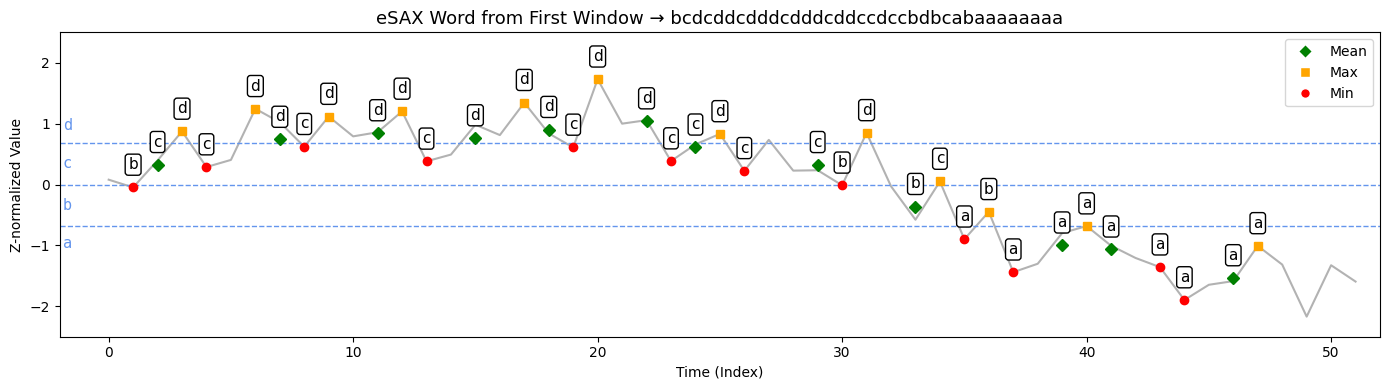}
    \caption{Example of eSAX word for the first window frame of \textit{programming} keyword}
    \label{fig:ESAX_WORD_EXAMPLE}
\end{figure}

The eSAX algorithm is similar to the SAX algorithm, with one adjustment: for each of the 12 segments, the mean, max, and min values are calculated. Each of these is then assigned a letter of the alphabet and these three letters are then ordered by their position in time within the segment (which value comes first, second, third). Figure~\ref{fig:ESAX_WORD_EXAMPLE} illustrates this process of how the first eSAX word is formed from the first sliding window. 

\subsubsection{K-means Clustering}
\begin{figure}[h!]
  \centering

  \begin{minipage}[b]{0.48\linewidth}
    \centering
    \includegraphics[width=\linewidth]{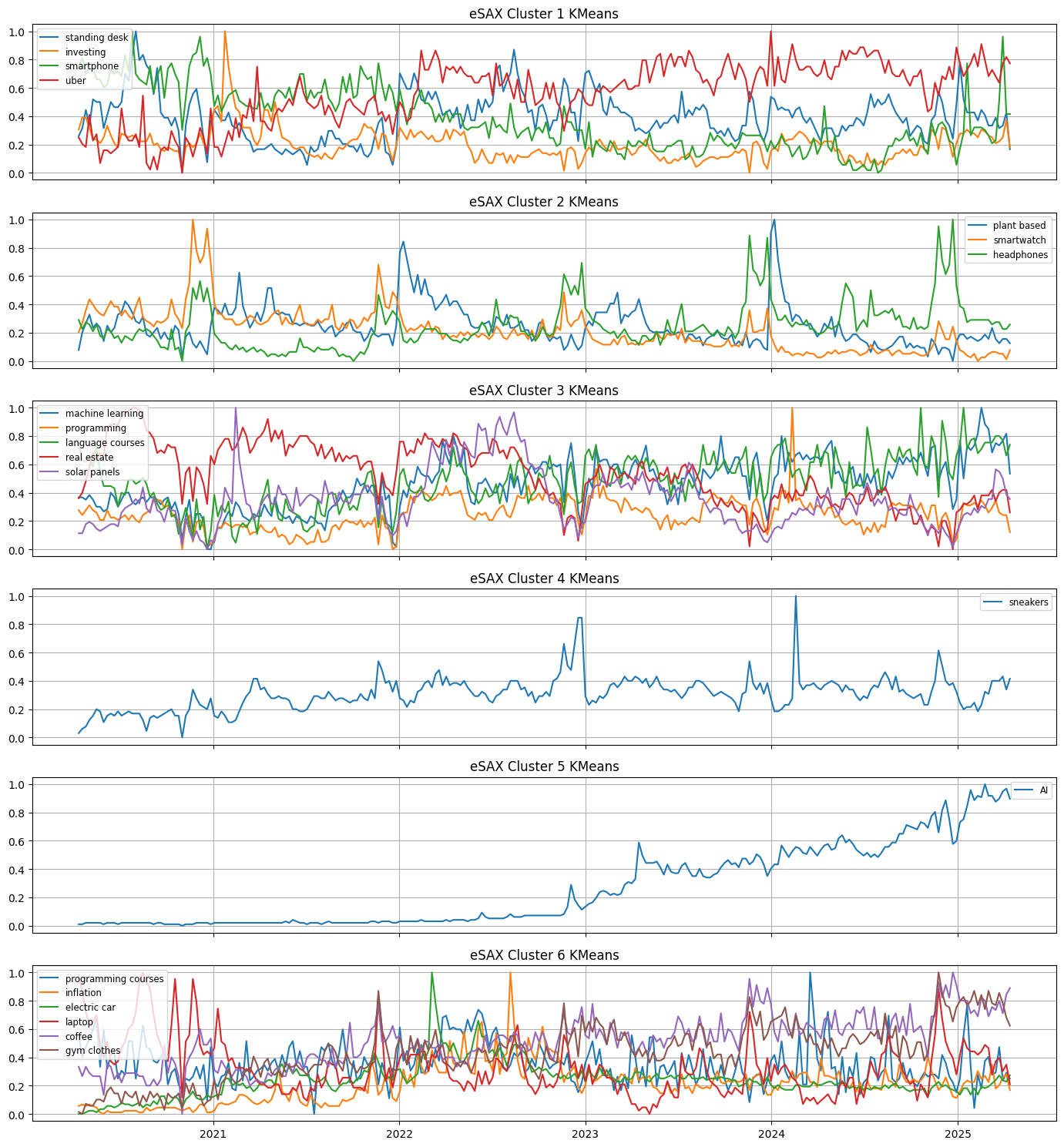}
    \caption{eSAX K-Means Clustering}
    \label{fig:11}
  \end{minipage}
  \hfill
  \begin{minipage}[b]{0.48\linewidth}
    \centering
    \includegraphics[width=\linewidth]{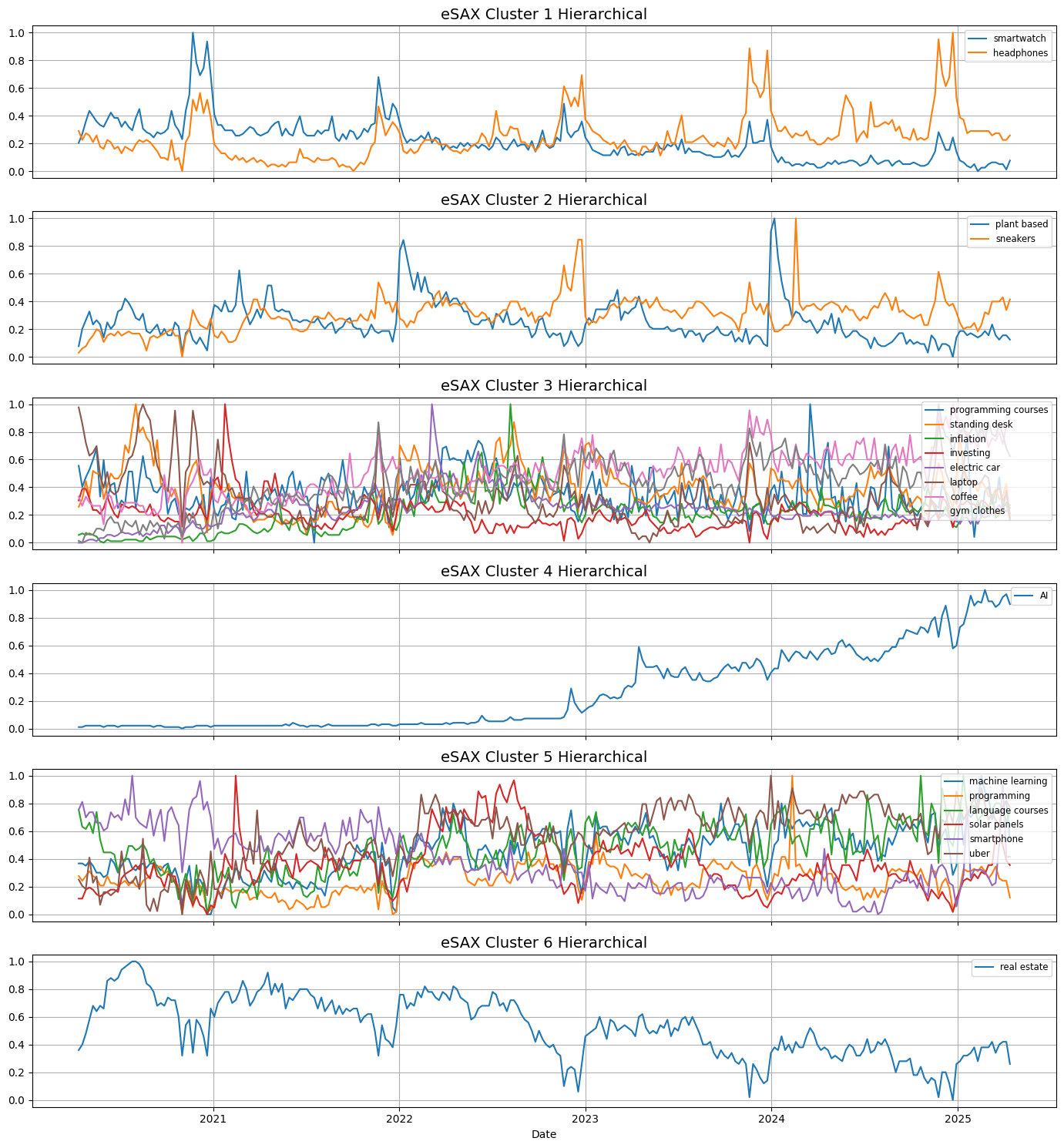}
    \caption{eSAX Hierarchical Clustering}
    \label{fig:12}
  \end{minipage}

\end{figure}

Despite having three times more information for clustering, the results shown in Figure~\ref{fig:11} are somewhat unexpected. The K-means algorithm appears to struggle with the increased input complexity, producing clusters that are less coherent than those obtained in the original SAX-based attempt. This is reflected in the silhouette score of 0.218 (see Table~\ref{tab:silhouette-scores}), which is notably lower than that of the SAX clustering. Clusters 1, 3, and 6 contain time series with irregular or chaotic behavior, while Clusters 4 and 5 each consist of only a single keyword. Cluster 2 is the most coherent grouping, capturing time series that follow a broadly similar trend.

\subsubsection{Hierarchical Clustering}
For the eSAX output, hierarchical clustering once again produced results that differed from those obtained using the original SAX representation. The resulting cluster structure is shown in Figure~\ref{fig:12}.

\smallbreak

Cluster 1 replicates a meaningful grouping already observed in the earlier SAX-based attempts, consisting of \textit{smartwatch} and \textit{headphones}. Notably, Clusters 4 and 6 each isolate a single keyword, \textit{AI} and \textit{real estate} respectively, suggesting that eSAX effectively captures the distinct shape of these time series. In contrast, Clusters 2 and 3 appear overloaded, containing numerous chaotic series that lack a clear or shared structure. These observations are supported both by visual inspection and by the silhouette score of 0.308, which reflects a modest clustering quality and aligns with the observed imbalance across clusters.

\subsection{SAX \& eSAX Comparison}
One consistent observation across the clustering methods is the recurring presence of at least one disorganized and poorly defined cluster. This cluster typically lacks internal coherence, showing no clear resemblance among its keywords or alignment with other clusters. This phenomenon is not unique to this study: similar findings have been reported in the literature. Hobbelhagen et al.~\cite{16}, for instance, noted that SAX-based clustering often produced one large, diffuse cluster alongside several smaller, well-separated ones.

\begin{table}[h!]
\centering
\caption{Silhouette Scores for SAX and eSAX Clustering}
\label{tab:silhouette-scores}
\begin{tabular}{L{4cm} C{4cm} C{4cm}}
\toprule
\textbf{Representation} & \textbf{KMeans Silhouette Score} & \textbf{Hierarchical Silhouette Score} \\
\midrule
SAX                     & 0.320                             & \textbf{0.355}                                  \\
eSAX                    & 0.218                             & 0.308                         \\
\bottomrule
\end{tabular}
\end{table}

Table~\ref{tab:silhouette-scores} presents the silhouette scores for each clustering method. Among them, SAX combined with hierarchical clustering achieves the highest score (0.355), indicating slightly better-defined clusters. However, all scores remain in the moderate range, reflecting the inherent complexity and variability of the data. As highlighted in Section~\ref{sec:EDA}, the selected time series are highly diverse in shape, trend, and volatility, making perfect clustering unlikely.

\smallbreak

For this reason, quantitative metrics such as silhouette scores were complemented by visual inspection. While some clusters appear fragmented, others consistently capture meaningful relationships. For example, \textit{smartwatch} and \textit{headphones} were frequently grouped together across different methods, reflecting their similar time series patterns.

\smallbreak

These results suggest that symbolic methods are well suited for capturing dominant, clearly defined trends in time series data. However, they tend to struggle with finer nuances and irregularities, even when enhanced representations like eSAX are used. In other words, adding minimum and maximum values to the average (as in eSAX) does not necessarily improve clustering performance when the underlying structure of the data is highly complex.

\smallbreak

Finally, it is important to emphasize that the purpose of applying SAX and eSAX in this study is not to identify the single best-performing model, but to explore how symbolic representations can be used to analyze time series data. This directly addresses the research question: how symbolic methods can be applied to understand consumer attention over time. The next section explores how topological methods, in particular, TDA, can offer a complementary and potentially more flexible perspective.

\section{Topological Methods}
\label{sec:tda}

In contrast to symbolic methods like SAX and eSAX, Topological Data Analysis (TDA) adopts a fundamentally different approach analyzing the shape of data using tools from algebraic topology. Instead of segmenting or transforming time series into symbolic representations, TDA captures structural features such as connected components, loops, and higher-dimensional voids, tracking how these features persist across multiple scales. This allows the method to uncover geometric and temporal patterns that may be obscured by noise or not easily detected using conventional techniques.

As discussed in Section~\ref{sec:TDA}, TDA is particularly useful for exploring the global structure of data. In this context, ``shape'' refers to the overall behavioral pattern of a time series, how interest rises, falls, or stabilizes over time. Unlike SAX-based methods, which analyze local trends within segments, TDA treats the time series as a whole, making it well-suited for complex, heterogeneous data like the consumer search trends studied here. Some time series exhibit seasonal or stable patterns, while others are more erratic or event-driven, complicating direct comparison. It is also worth mentioning, that unlike SAX-based methods, TDA does not rely on assumptions about the underlying distribution of the data. Hence, the original (unstandardized) time series are used for TDA, as additional preprocessing steps (such as normalization) could distort the intrinsic geometry that TDA aims to capture.

TDA addresses this challenge by forgoing assumptions about stationarity or distribution, instead extracting structural features that persist across scales. This is achieved through the framework of persistent homology, which produces a multiscale topological summary of each time series. These summaries are then converted into vectorized representations, specifically, persistence landscapes, that can be used for clustering based on shape alone. Each step of this process is described in the following subsections.

\subsection{Overview of the TDA Pipeline}
\label{subsec:tda-overview}

The Topological Data Analysis (TDA) pipeline used in this study is designed to extract structural features from time series that reflect their overall shape and temporal dynamics. Unlike traditional methods that rely on local values or statistical assumptions, TDA captures topological properties, such as connected components and loops, that persist across multiple scales of the data.

The process begins by transforming each time series into a high-dimensional point cloud using a {\it sliding window embedding} (see \S~\ref{sec:slidingwindow}). This embedding maps short consecutive segments of the time series into vectors, effectively unfolding temporal patterns into geometric space. From this point cloud, a Vietoris--Rips complex is constructed to model the connectivity between points based on pairwise distances.

{\it Persistent homology} is then applied to this complex to identify topological features that appear (are ``born'') and disappear (``die'') as the scale of analysis changes (see (see \S~\ref{hom})). These features are summarized in {\it persistence diagrams}, which record the lifespan of each topological feature (see \S~\ref{sec:per_dia}).

To enable clustering and further analysis, the persistence diagrams are converted into vectorized representations known as {\it persistence landscapes}. These landscapes capture the most prominent topological features in a format suitable for machine learning techniques.

Finally, clustering is applied to the persistence landscapes to identify groups of time series that share similar topological characteristics. Each step of this pipeline is described in detail in the subsections that follow.


\subsection{Sliding Window Embedding}
\label{sec:slidingwindow}

To implement TDA, the raw time series data must first be transformed into a suitable format. This is achieved through \textit{Sliding Window Embedding}, a technique that converts each time series into a \textit{point cloud}, that is, a set of data points in a high-dimensional space that captures local temporal patterns.

This embedding is defined by two key parameters: the \textit{embedding dimension} \(d\) and the \textit{time delay} \(\tau\). While there is no universally optimal choice, these parameters significantly influence the geometry of the resulting point cloud. Inappropriate values can lead to overly flat or sparse structures, which in turn affect the accuracy of subsequent topological analysis.

In many TDA applications, an embedding dimension of \(d = 10\) and a time delay of \(\tau = 1\) to \(3\) are commonly used. For instance, Truong (2017)~\cite{29} applies embedding with \(d = 10\) and a time delay of \(\tau = 1\) to financial time series, supporting this choice with analysis of autocorrelation and false nearest neighbours. However, for the weekly Google Trends data used in this study, different parameter combinations were tested to find a configuration that balances resolution, interpretability, and computational efficiency. The final choice was an embedding dimension of \(d = 6\) and a time delay of \(\tau = 3\). This setup captures local fluctuations across short periods (i.e., three weeks), which is important for identifying short-term behavioral patterns in consumer attention, while avoiding over-fitting to noise.

Although somewhat heuristic, this parameter choice proved robust throughout the analysis. It yielded comparable or slightly better results than the more typical configuration (e.g., \(d = 10\), \(\tau = 1\)), particularly in terms of the clarity of persistence diagrams, barcodes, and clustering outputs. Moreover, it reduced computational complexity without compromising the overall quality of the results.

\smallbreak

Once the time series are embedded, each is represented as a point cloud in \( \mathbb{R}^6 \). The next step is to construct topological structures on these point clouds to capture their shape across scales. This is described in the following subsection.


\subsection{Vietoris–Rips Complexes and Persistent Homology}
\label{hom}

\begin{figure}[h!]
    \centering
    \includegraphics[width=1\linewidth]{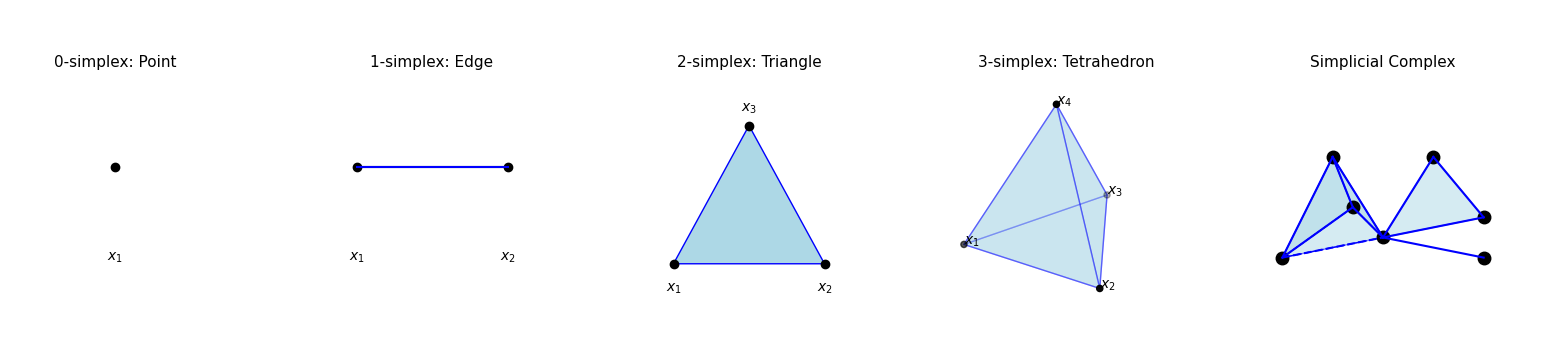}
    \caption{Simplices and a Simplicial Complex Example}
    \label{fig:13}
\end{figure}

Once the sliding window embeddings transform the time series into point clouds, we build \textit{simplicial complexes} to capture their shape. A simplicial complex is a geometric structure made of simplices, i.e. points (0-simplices), edges (1-simplices), triangles (2-simplices), tetrahedra (3-simplices), and their higher-dimensional analogues. This is illustrated in Figure~\ref{fig:13}.

\begin{figure}[h!]
    \centering
    \includegraphics[width=1\linewidth]{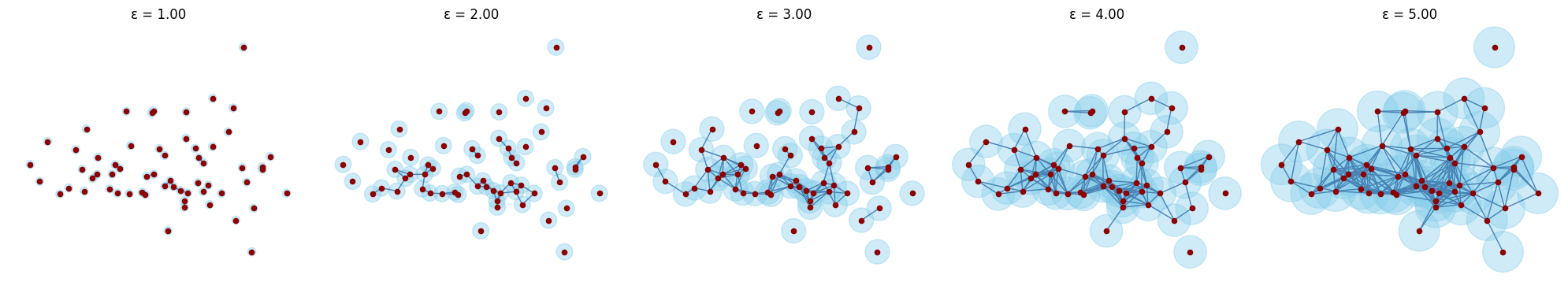}
    \caption{The Process of Building Simplicial Complexes}
    \label{fig:14}
\end{figure}

To construct such complexes from a point cloud, we imagine placing a ball of radius \( r \) around each point. As \( r \) increases, the balls begin to intersect. When two balls overlap, we form an edge; when three intersect pairwise, we form a triangle, and so on. The resulting structure encodes the way data points cluster, connect, and fill higher-dimensional spaces. Figure~\ref{fig:14} demonstrates this idea.

A widely used type of simplicial complex is the \textit{Vietoris–Rips complex}, which we use in this study. It connects points that lie within a given distance threshold and builds higher-dimensional simplices accordingly. By varying this threshold, we track how the topological structure of the dataset evolves across multiple scales.

\subsection{Persistence Diagrams}
\label{sec:per_dia}

As we increase the scale parameter \( r \), topological features such as connected components, loops, and voids begin to emerge and disappear. \textit{Persistent homology} is the method used to track the appearance and disappearance of these features over scale.

\begin{figure}[h!]
    \centering
    \includegraphics[width=0.4\linewidth]{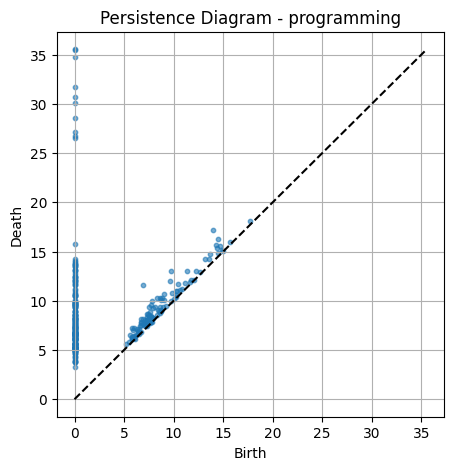}
    \caption{Persistence Diagram of the keyword 'programming'}
    \label{fig:20}
\end{figure}

Each topological feature is assigned a \textit{birth time} (when it first appears) and a \textit{death time} (when it is filled in or merges with another structure). These are visualized in a \textit{persistence diagram}, where each point \((b, d)\) corresponds to a feature born at scale \( b \) and dying at scale \( d \). An example is shown in Figure~\ref{fig:20}.

Points near the diagonal represent short-lived (low-persistence) features, which are typically considered noise. In contrast, points farther from the diagonal indicate more persistent and meaningful structures, those that capture the dominant shape of the data. This makes the persistence diagram a compact summary of the dataset’s most significant topological features.


\subsection{Persistent Barcodes}
\textit{Persistence barcodes} are an alternative visualization of the same information contained in persistence diagrams. Both tools track topological features across scales, but while persistence diagrams display each feature as a point with coordinates (birth, death), barcodes represent features as horizontal bars that span their lifespan over a filtration scale.

This perspective provides a more intuitive way to follow the entire topological evolution of the dataset. Rather than focusing only on the most persistent features, persistence barcodes make it easier to observe when features appear and disappear during the construction of the simplicial complex.

\begin{figure}[h!]
    \centering
    \includegraphics[width=0.70\linewidth]{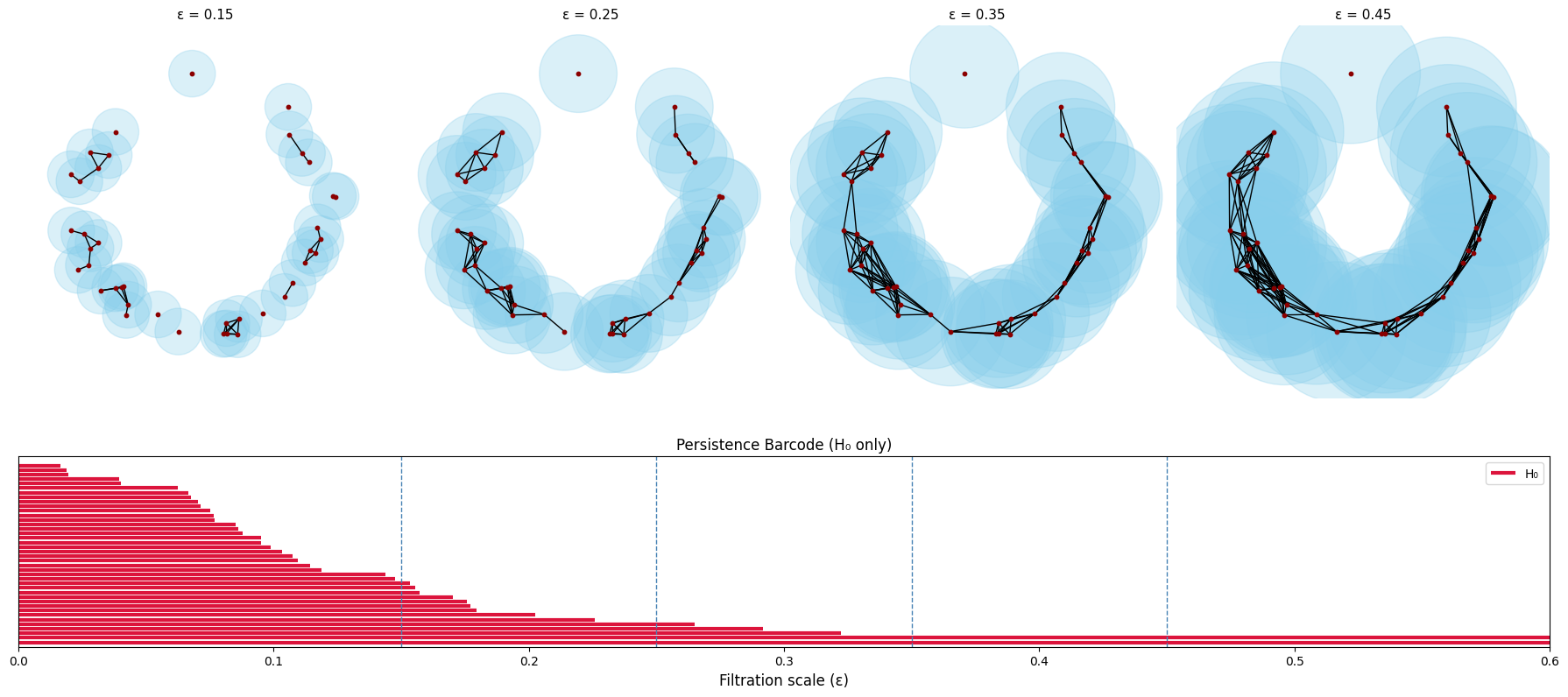}
    \caption{The Process of Building a Persistence Barcode}
    \label{fig:15}
\end{figure}

Figure~\ref{fig:15} illustrates this process on synthetically generated data. The x-axis corresponds to the \textit{filtration value}, that is, a scale parameter indicating the size of the growing balls around each point in the cloud. Each horizontal bar represents a topological feature: it begins at its birth and ends at its death. The longer the bar, the more persistent the feature, and thus, the more significant it is.

\begin{figure}[h!]
    \centering
    \includegraphics[width=0.5\linewidth]{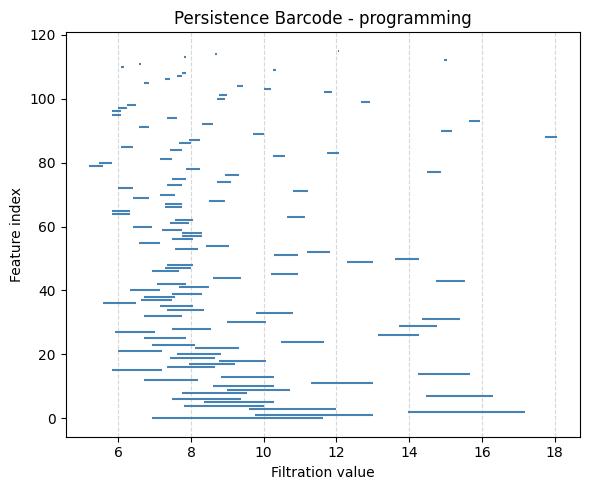}
    \caption{Persistence Barcode of 'programming' Keyword}
    \label{fig:21}
\end{figure}

Figure~\ref{fig:21} presents a real barcode computed for the \textit{programming} keyword. Most features are short-lived, appearing and disappearing between filtration values 6 and 10. Only a few features persist beyond filtration 12, indicating that the point cloud becomes highly connected at early scales. The longest bars, shown at the bottom, correspond to the most significant structural features in the data.

In summary, persistence diagrams and persistence barcodes provide two equivalent representations of persistent homology. While both are valuable tools for visual inspection and topological interpretation, they are not directly usable in statistical or machine learning models. To bridge this gap, persistence landscapes are introduced in the next subsection as a vectorized format suitable for downstream analysis.

\subsection{Persistent Landscapes}
\textit{Persistent landscapes} convert persistence barcodes into structured, vectorized summaries that preserve key topological information. Unlike barcodes or diagrams, persistence landscapes produce numerical representations that can be directly used in statistical models and machine learning algorithms.

\smallbreak

Each bar in the barcode is transformed into a triangle-shaped function called a \textit{tent function}. This function peaks at the midpoint between the birth and death of the feature, and it tapers to zero at both ends. In this way, each triangle encodes the ``lifespan'' of a topological feature. When multiple features are present, their triangle functions overlap across the filtration axis, forming a layered structure. At each point along this axis, we evaluate all active tent functions and sort their values from highest to lowest. 

\smallbreak
Formally, the $k$-th persistence landscape function $\lambda_k(t)$ is defined as the $k$-th largest value among all tent functions at point $t$:
\[
\lambda_k(t) = \text{$k$-th largest value of } \{f_i(t)\}_{i=1}^n,
\]
where each $f_i(t)$ is a tent function constructed from a persistence interval $[b_i, d_i]$ with peak at $(b_i + d_i)/2$ and height $(d_i - b_i)/2$. The landscape $\lambda_1$ captures the most persistent topological feature at each $t$, $\lambda_2$ the second most persistent, and so on.

\begin{figure}[h!]
    \centering
    \includegraphics[width=1\linewidth]{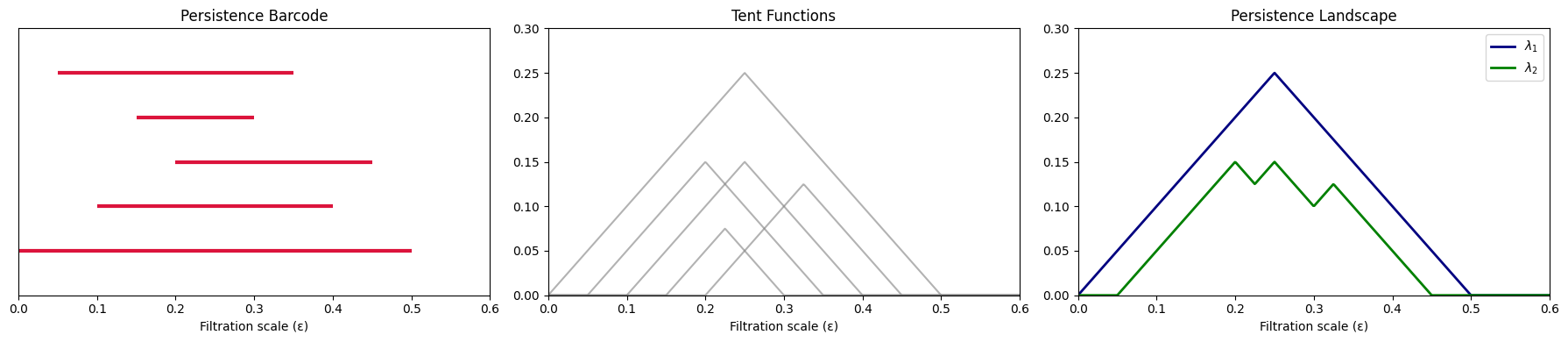}
    \caption{Persistence Landscape Process}
    \label{fig:16}
\end{figure}

The first landscape layer records the highest value at each point (i.e. the most persistent feature active there), the second layer records the second-highest value, and so on. The longer a feature persists, the taller and wider its triangle, making it more influential in the landscape. This process is illustrated in Figure~\ref{fig:16} using simple synthetic data. 

\begin{figure}[h!]
  \centering

  \begin{minipage}[b]{0.48\linewidth}
    \centering
    \includegraphics[width=\linewidth]{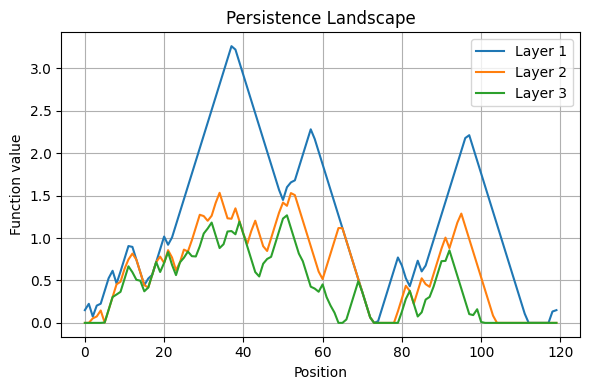}
    \caption{Persistence Landscape of All H1 Features for ``programming''}
    \label{fig:23}
  \end{minipage}
  \hfill
  \begin{minipage}[b]{0.48\linewidth}
    \centering
    \includegraphics[width=\linewidth]{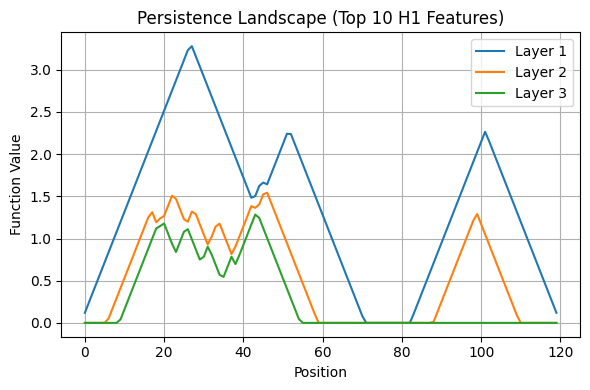}
    \caption{Persistence Landscape of Top 10 H1 Features for ``programming''}
    \label{fig:24}
  \end{minipage}

\end{figure}

Figures~\ref{fig:23} and~\ref{fig:24} illustrate how these tent-shaped functions behave on real data. Figure~\ref{fig:23} shows the full persistence landscape of the \textit{programming} keyword. The three visible layers correspond to the top three most persistent topological features at each scale. Figure~\ref{fig:24} simplifies this view by isolating only the top 10 most persistent features, making the shape of the data easier to interpret.

\smallbreak

To use this information for clustering, each landscape is discretized into a numerical vector. All such vectors (one per keyword) are then stacked into a matrix, where each row corresponds to a keyword and each column represents a point on the filtration axis. This matrix serves as the input for the clustering analysis in the following subsection.

\subsection{TDA Clustering}
Having transformed each time series into a persistence landscape vector, a numeric summary of its underlying shape, we now apply clustering techniques based on topological features. The goal is to explore whether TDA can provide meaningful groupings that reflect structural similarity in consumer attention patterns.

\smallbreak

Two strategies are tested to illustrate the flexibility of TDA. The first applies K-Means clustering to persistence landscapes computed solely from the most persistent loops (dimension 1 features). The second combines connected components and loops (dimensions 0 and 1), filtered for persistence, and uses Hierarchical Clustering. These strategies are not meant to be directly compared with each other or with SAX-based clustering, but rather to highlight the range of representations and decisions TDA enables.


\subsubsection{K-Means Clustering}
The clusters obtained from applying K-Means to the persistence landscapes are shown in Figure~\ref{fig:25}. From a visual standpoint, the clustering results are promising: TDA appears to resolve some of the challenges observed with SAX and eSAX. The time series are now distributed more evenly, and each cluster captures distinct, meaningful variations in shape and trend.

\smallbreak

Unlike earlier approaches that produced a few tightly grouped clusters alongside one or two ``chaotic'' catch-all clusters, the TDA-based clustering presents more balanced groupings. Notably, even structurally complex time series, such as those for \textit{AI}, \textit{inflation}, and \textit{solar panels}, are handled more effectively, although they still exhibit some clustering difficulty due to their unique behaviors.

\smallbreak

Despite these qualitative strengths, the evaluation metric tells a different story. The silhouette score for K-Means is only 0.146, indicating weak boundary separation and limited internal coherence within clusters. This reveals a key challenge in evaluating TDA: visual and structural insights may not always align with conventional performance metrics.

\subsubsection{Hierarchical Clustering}
\begin{figure}[h!]
  \centering
  \begin{minipage}[b]{0.48\linewidth}
    \centering
    \includegraphics[width=\linewidth]{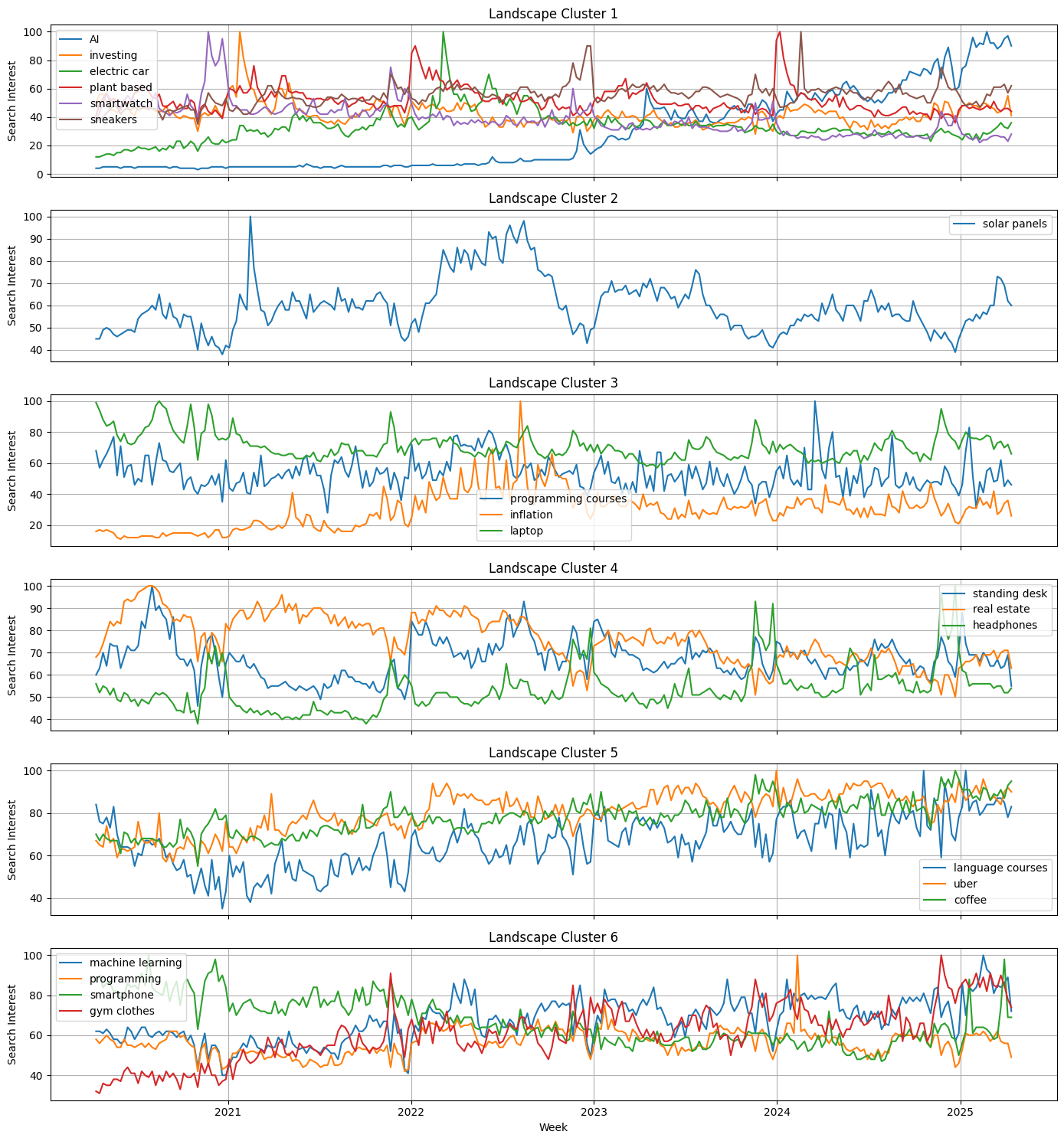}
    \caption{TDA K-Means Clustering}
    \label{fig:25}
  \end{minipage}
  \hfill
  \begin{minipage}[b]{0.48\linewidth}
    \centering
    \includegraphics[width=\linewidth]{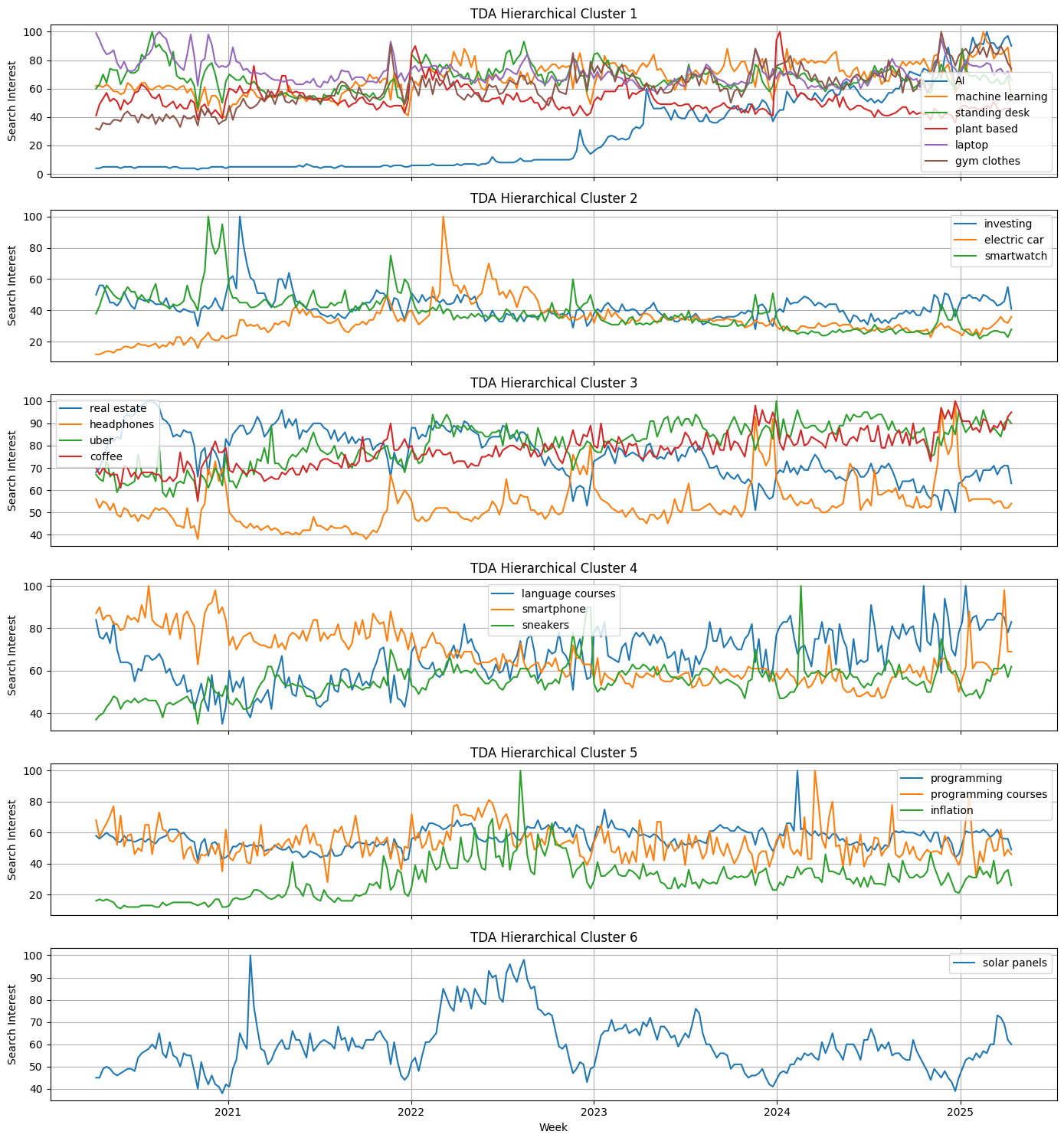}
    \caption{TDA Hierarchical Clustering}
    \label{fig:26}
  \end{minipage}
\end{figure}

To explore whether this performance can be improved, we apply Hierarchical Clustering using a broader set of topological features, specifically, both connected components and loops filtered for persistence. The results are illustrated in Figure~\ref{fig:26}.

\smallbreak

This approach leads to a significant increase in silhouette score, reaching 0.375, the highest across all clustering methods evaluated. However, the visual clarity of the clusters is reduced compared to the K-Means result. The clusters appear less distinct, and some exhibit internal heterogeneity. Still, this trade-off demonstrates the versatility of TDA: the method can be tuned to prioritize either interpretability or statistical validation, depending on the context and application.

\smallbreak

In summary, the TDA results underscore the central motivation of this study: to investigate whether topological techniques can offer a richer, structure-aware alternative to symbolic time series analysis. Rather than simplifying complexity, TDA uses and interprets it, extracting meaningful, multiscale patterns from raw data without enforcing strict statistical assumptions.

\section{Discussion \& Conclusions}
\label{sec:discussion}

This section reflects on the comparative performance, interpretability, and limitations of the three methods explored in this paper: SAX, eSAX, and TDA. We summarize the findings, assess the strengths and weaknesses of each method, and suggest directions for future research.

\subsection{Method Comparison and Interpretation}

Symbolic methods such as SAX and eSAX offer a simple, interpretable, and computationally efficient approach to time series clustering. By assigning discrete symbols to segments of the time series, these methods can rapidly group sequences with similar trends. In our study, both SAX and eSAX produced promising clustering results, grouping several time series with comparable behavior. However, both methods consistently yielded one cluster that lacked interpretability, a ``catch-all'' group for irregular or ambiguous series. These clusters, although not incorrect, reflect a limitation of symbolic approaches: their tendency to oversimplify complex or volatile data.

Topological Data Analysis (TDA), on the other hand, adopts a fundamentally different methodology. Rather than discretizing or summarizing, it analyzes the shape and structure of the data in a continuous and multiscale fashion. Persistence landscapes derived from TDA enabled clustering that better captured underlying patterns, especially for irregular or noisy time series. Most notably, TDA avoided the emergence of an unstructured ``everything else'' cluster. The flexibility of TDA was demonstrated by two approaches: the first using only persistent loops, which resulted in visually meaningful clusters but lower silhouette scores, and the second combining loops and connected components, which achieved the highest silhouette score (0.375) across all models.

Both symbolic and topological methods have their strengths and weaknesses, and each method approaches the time series in a different way. SAX and eSAX work by assigning symbols to the data, which is fast and interpretable but limited in depth. TDA looks at the data from a more complex perspective, which is expensive to compute and not easily interpretable. 

\smallbreak

In addition to the silhouette score, we also report the {\it Davies–Bouldin Index} (DBI)~\cite{30} for further evaluation of clustering performance. DBI is another commonly used internal evaluation metric for clustering. It measures the average similarity between each cluster and its most similar counterpart, based on the ratio of within-cluster dispersion to between-cluster separation. Lower DBI values indicate better clustering performance, with well-separated and compact clusters generally producing lower scores. The DBI score for SAX-based K-means clustering is 0.749 (see Table~\ref{tab:final-model-comparison}), which reinforces the interpretation that the clusters are moderately separated but not clearly defined. This result aligns with the silhouette score in suggesting that while SAX captures some meaningful structure, it struggles to distinguish more complex patterns.

The Silhouette Score indicates that TDA yields better-separated and more well-defined clusters, with higher values suggesting improved between-cluster separation and within-cluster cohesion. In contrast, SAX achieves a lower DBI, reflecting tighter and more compact clusters.
Overall, despite SAX producing more compact groupings, the higher Silhouette Score suggests that TDA is more effective for clustering in general, particularly when cluster separation and structural interpretability are important. This aligns with the intuition that topological representations capture global shape features that facilitate more meaningful distinctions between clusters.

\begin{table}[h!]
\centering
\caption{Comparison of Clustering Evaluation Scores Across All Methods}
\label{tab:final-model-comparison}
\begin{tabular}{C{2.2cm} L{4.2cm} C{2.2cm} C{2.2cm} C{2.2cm}}
\toprule
& \textbf{Metric} & \textbf{SAX} & \textbf{eSAX} & \textbf{TDA} \\
\midrule
& Silhouette Score         & 0.320 & 0.218 & 0.146 \\
K-Means & &&&\\
& Davies–Bouldin Index     & 0.749 & 0.777 & 1.133 \\
\midrule
& Silhouette Score   & 0.355 & 0.308 & \textbf{0.375} \\
Hierarchical & &&&\\
& Davies–Bouldin Index & {\bf 0.618} & 0.652 & 0.723 \\
\bottomrule
\end{tabular}
\end{table}

The trade-off between silhouette score and Davies--Bouldin Index (DBI) highlights a nuanced finding. While SAX produced more compact clusters (lower DBI), TDA resulted in better-separated and more distinct clusters (higher silhouette score). These metrics together suggest that TDA is more effective when structural differentiation and interpretability are priorities, even if the clusters appear less compact.

A particularly important finding is TDA's ability to meaningfully group ``difficult'' keywords such as \textit{AI} and \textit{inflation}. These time series were treated as outliers by symbolic methods, but TDA was able to assign them to clusters based on topological features. Moreover, TDA proved to be highly adaptable: by adjusting parameters and selecting specific topological features, we could shift the balance between visual clarity and numerical performance.

\subsection{Limitations and Scope}\label{lim}

While this study offers meaningful insights into the comparative performance of symbolic and topological time-series clustering methods, certain limitations should be acknowledged.

First, the evaluation primarily relied on internal clustering metrics such as silhouette score and Davies–Bouldin index. These provide useful quantitative benchmarks but do not fully capture domain-specific interpretability or practical marketing value. Future work could incorporate external validation strategies, such as expert annotations or labeled behavioral datasets.

Second, the methods employed were tested with representative, but not fully optimized, hyperparameter settings. This was a deliberate choice to maintain methodological transparency and reduce over-fitting on a single dataset. Nonetheless, a more exhaustive search across embedding parameters or SAX breakpoints could reveal further performance potential.

Third, Topological Data Analysis, while offering powerful interpretability and robustness, entails higher computational complexity compared to symbolic approaches. This may present challenges in large-scale or low-latency environments. However, recent advances in fast persistent homology and dimensionality reduction pipelines are making real-time applications increasingly feasible (see for example \cite{31}).

Overall, these limitations reflect the scope and objectives of an exploratory comparative study rather than methodological shortcomings. They also highlight opportunities for future research discussed below.

\subsection{Future Research}\label{future}

It is worth mentioning that the insights
offered by TDA may go beyond clustering. Future work may explore several promising directions.

\subsubsection{Hybrid Symbolic–Topological Pipelines}
One is the development of hybrid methods that combine symbolic preprocessing with topological analysis. For example, SAX could be used to reduce noise or detect general trends, followed by TDA to extract structural nuances. Another direction involves expanding the evaluation to include external validation, perhaps with labeled datasets or domain expert feedback.

\subsubsection{Systematic Parameter Optimization}
There is also scope for improving parameter selection. More systematic exploration of embedding dimension, delay, and filtration thresholds may yield better performance. Additionally, the interpretability of TDA results could be enhanced through new visualizations or mapping topological features back to specific time series segments.

\subsubsection{Real-Time TDA Pipelines for Streaming Data}
A promising extension of this work would involve developing a real-time pipeline where Google Trends data are streamed, embedded, and analyzed using online persistent homology algorithms. This would enable the dynamic monitoring of consumer behavior, with alerts triggered by topological shifts.

\subsubsection{TDA-Informed Predictive Modeling}
Further research could explore the integration of TDA-derived features (e.g., Betti curves, lifespan distributions, topological entropy) into predictive models for trend forecasting. Applications in anomaly detection, event anticipation, and behavioral segmentation based on topological signatures may offer new tools for interpretable and adaptive marketing analytics.

\subsubsection{Implications for Marketing Analytics}
Topological Data Analysis provides a robust and interpretable lens into the evolving shape of consumer interest over time. When applied to streaming search data, persistent homology computed over sliding windows can reveal structural changes in consumer behavior that are often missed by traditional volume-based models. For instance, the sudden appearance or disappearance of topological features such as loops or connected components may indicate the emergence of new trends, responses to external shocks, or anomalies in public sentiment. By comparing recent and historical topological summaries, one may construct a \emph{topological anomaly score}, namely, a quantitative signal that could trigger timely marketing actions, including targeted campaigns or crisis management. A key advantage of TDA in this context is its robustness to noise and invariance to scale, which makes it particularly suited to irregular and high-variance data. Moreover, because of its coordinate-free nature, TDA captures shape-based similarities even when the timing or amplitude of consumer responses varies across categories. Persistent topological patterns may further reveal latent periodicity, offering a principled foundation for seasonality-aware marketing strategies.

\subsection{Concluding Remarks}\label{conclusion}

In summary, SAX and eSAX proved effective for time series with stable, repetitive patterns, offering both computational efficiency and interpretability. However, their symbolic simplification came at the cost of failing to capture complex variations. In contrast, TDA, though computationally heavier and less intuitive, demonstrated greater capacity to capture global structural features and handle irregular behavior. The integration of symbolic and topological approaches therefore appears most promising.

\smallbreak

We conclude that understanding the evolution of consumer attention over time requires analytical tools that move beyond local patterns and summary trends. Topological methods, despite their mathematical complexity, offer a flexible and powerful framework for revealing deeper, shape-based structures in behavioral data, insights that are often inaccessible to traditional time-series analysis.

\bibliographystyle{plainurl}
\bibliography{references}




\end{document}